\documentclass[10pt,journal,compsoc]{IEEEtran}
\ifCLASSOPTIONcompsoc
  \usepackage[nocompress]{cite}
\else
  \usepackage{cite}
\fi
\ifCLASSINFOpdf
\else
\fi
\usepackage{ragged2e} 
\usepackage[table]{xcolor}
\usepackage{graphicx}
\usepackage{amssymb}
\usepackage{amsmath}
\usepackage{multirow}
\usepackage{tabularx}
\usepackage{balance}
\usepackage{dsfont}
\usepackage{url}
\usepackage{color}
\usepackage{amsthm}
\usepackage[ruled,linesnumbered]{algorithm2e}
\usepackage[export]{adjustbox}
\usepackage{subcaption}
\usepackage[font=footnotesize,labelfont=bf]{caption}
\usepackage{longtable}
\usepackage{tabu}
\usepackage{diagbox,booktabs}
\usepackage{multicol}
\usepackage{makecell}
\usepackage{enumitem}

\usepackage{overpic}
\definecolor{citecolor}{RGB}{65,105,225}
\usepackage[breaklinks=true,colorlinks,citecolor=citecolor,bookmarks=false]{hyperref}

\definecolor{dg}{rgb}{0,0.694,0.298}
\definecolor{purple}{rgb}{0.4,0.176,0.569}
\definecolor{royalblue}{RGB}{65,105,225}
\usepackage{pifont}
\newcommand{\reqref}[1]{Eq.~\eqref{#1}}

\makeatletter
\DeclareRobustCommand\onedot{\futurelet\@let@token\@onedot}
\def\@onedot{\ifx\@let@token.\else.\null\fi\xspace}
\def\eg{\emph{e.g}\onedot} 
\def\ie{\emph{i.e}\onedot}

\makeatother

\definecolor{americanrose}{rgb}{1.0, 0.01, 0.24}

\def\ie{\textit{i.e.}}
\def\eg{\textit{e.g.}}

\definecolor{mygray}{gray}{.9}
\definecolor{mypink}{rgb}{.99,.91,.95}
\definecolor{mycyan}{cmyk}{.3,0,0,0}
\definecolor{myyellow}{rgb}{.99,.9,0}

\begin{document}
%
\title{On the Robustness of Segment Anything}
%
%
%
%

\author{Yihao Huang*,~
        Yue Cao*,~
        Tianlin Li$^\dag$,~
        Felix Juefei-Xu,~\IEEEmembership{Member,~IEEE,}~
        Di Lin,~\IEEEmembership{Member,~IEEE,}~\\
        Ivor W.~Tsang,~\IEEEmembership{Fellow,~IEEE,}~
        Yang Liu,~\IEEEmembership{Senior Member,~IEEE,}~
        Qing Guo$^\dag$,~\IEEEmembership{Member,~IEEE}
\IEEEcompsocitemizethanks{\IEEEcompsocthanksitem Yihao Huang, Yue Cao, Tianlin Li, and Yang Liu are with Nanyang Technological University, Singapore
\IEEEcompsocthanksitem Felix Juefei-Xu is with New York University, New York, NY 10012, USA
\IEEEcompsocthanksitem Di Lin is with the College of Intelligence and Computing, Tianjin University, China.
\IEEEcompsocthanksitem Ivor W. Tsang and Qing Guo are with the Centre for Frontier AI Research (CFAR), A*STAR, Singapore, and Institute of High Performance Computing (IHPC), A*STAR, Singapore.
\IEEEcompsocthanksitem *~Yihao Huang and Yue Cao are the co-first authors 
\IEEEcompsocthanksitem $^\dag$~Qing Guo and Tianlin Li are the corresponding authors (tsingqguo@ieee.org).
}

\thanks{Manuscript}}

\markboth{Preprint}%
{Shell \MakeLowercase{\textit{et al.}}: Bare Demo of IEEEtran.cls for Computer Society Journals}
%



\IEEEtitleabstractindextext{%
\begin{abstract}
Segment anything model (SAM) has presented impressive objectness identification capability with the idea of prompt learning and a new collected large-scale dataset. Given a prompt (\eg, points, bounding boxes, or masks) and an input image, SAM is able to generate valid segment masks for all objects indicated by the prompts, presenting high generalization across diverse scenarios and being a general method for zero-shot transfer to downstream vision tasks. Nevertheless, it remains unclear whether SAM may introduce errors in certain threatening scenarios. Clarifying this is of significant importance for applications that require robustness, such as autonomous vehicles. In this paper, we aim to study the testing-time robustness of SAM under adversarial scenarios and common corruptions. To this end, we first build a testing-time robustness evaluation benchmark for SAM by integrating existing public datasets. Second, we extend representative adversarial attacks against SAM and study the influence of different prompts on robustness. Third, we study the robustness of SAM under diverse corruption types by evaluating SAM on corrupted datasets with different prompts. With experiments conducted on SA-1B and KITTI datasets, we find that SAM exhibits remarkable robustness against various corruptions, except for blur-related corruption. Furthermore, SAM remains susceptible to adversarial attacks, particularly when subjected to PGD and BIM attacks. We think such a comprehensive study could highlight the importance of the robustness issues of SAM and trigger a series of new tasks for SAM as well as downstream vision tasks.
\end{abstract}

}

\maketitle

\IEEEdisplaynontitleabstractindextext

\IEEEpeerreviewmaketitle

\IEEEraisesectionheading{\section{Introduction}
\label{sec:introduction}}
With the emergence of large language models (\eg, ChatGPT \cite{openai2023gpt4}), it has been demonstrated that training these models on vast amounts of data can effectively enable them to learn common tasks. In the field of computer vision, the Segment Anything Model (SAM) \cite{kirillov2023segany} is the pioneering foundation model that adopts this paradigm. As a result, SAM exhibits remarkable generalization capabilities when applied to image segmentation tasks.

SAM is a highly versatile model that has been trained on a comprehensive dataset (SA-1B), comprising over 1 billion masks derived from 11 million images. By utilizing prompt inputs such as points or bounding boxes, SAM demonstrates exceptional performance in zero-shot transfer learning for downstream segmentation tasks, even with new image distributions. Subsequent studies have assessed its generalization capabilities across diverse domains and scenarios, including medical images \cite{he2023accuracy,wu2023medical,zhang2023customized,roy2023sam,hu2023skinsam,li2023polyp}, camouflaged object detection \cite{tang2023can}, salient object segmentation \cite{ji2023segment}. Furthermore, SAM is also used as a foundation model for promoting other computer vision tasks such as semantic segmentation \cite{lin2018multi,lin2019zigzagnet,lin2022tagnet,shen2020ranet}, image editing \cite{yu2023inpaint,xie2023edit}, style transfer \cite{liu2023any}, object counting \cite{ma2023can}. Nevertheless, considering SAM's broad application scope, ensuring robustness becomes a paramount concern. 
A few notable works \cite{wang2023empirical, zhang2023attack} have explored the evaluation of SAM's robustness against corruption and adversarial attacks respectively. Compared with them, we not only employ the evaluation on SA-1B and KITTI datasets, \ie, examining the effectiveness of SAM under both identically distributed (IID) and out-of-distribution (OOD) conditions, but also discuss SAM's robustness in a more fine-grained way by evaluating SAM on relatively small and big objects. Furthermore, we additionally explore attacks specifically designed for segmentation tasks rather than classification tasks, which is more reasonable than that in \cite{zhang2023attack}. In experiments, we provide more detailed quantitative results than previous work, which can provide comprehensive information for a better understanding of the results.

Our primary objective is to investigate the test-timing robustness of SAM in the face of adversarial scenarios intentionally created by humans, as well as common corruptions that occur naturally. To accomplish this, we establish an evaluation benchmark using publicly available datasets, specifically SA-1B and KITTI \cite{Alhaija2018IJCV}. The experiments conducted on the SA-1B dataset aim to assess SAM's robustness in an independently and identically distributed (IID) setting. Conversely, the experiments carried out on the KITTI focus on evaluating SAM's robustness in safety-critical applications under out-of-distribution (OOD) conditions. Additionally, we perform experiments to evaluate SAM's robustness concerning relatively small or large objects within the KITTI, which allows for a more detailed and comprehensive analysis. Surprisingly, our findings indicate that SAM exhibits remarkable robustness against various corruptions, except for blur-related corruption. Furthermore, our research reveals that SAM remains susceptible to adversarial attacks, particularly when subjected to PGD and BIM attacks.

\section{Robustness of Segment Anything Model}

In this section, we first introduce the segment anything model (SAM) \cite{kirillov2023segany}. Then, we design the adversarial attack methods against the SAM. Meanwhile, we detail the common corruption against SAM.

\subsection{Segment Anything Model}

Foundation models have presented impressive zero-shot and few-shot generalizations \cite{brown2020language,wiggins2022opportunities}. Such progresses have been extended for training multi-modal fundamental models like CLIP \cite{radford2021learning}. Recently, Meta further extends the foundation model training to the vision domain and develops the SAM \cite{kirillov2023segany}, which is to build a foundation model for image segmentation. Specifically, SAM develops a promotable segmentation task to predict an object mask for a given segmentation prompt. The prompt could be a point, a bounding box, a mask, or a text. The promotable segmentation task can be used to solve general downstream segmentation tasks like zero-shot single-point valid mask evaluation, zero-shot edge detection, zero-shot object proposal, and zero-shot instance segmentation. In general, we can formulate the SAM in the following way
%
\begin{align}
    \mathcal{M} = \text{SAM}(\mathbf{I},\text{P}), \text{s.t.}~\text{P}\in\mathcal{P}=\{\text{P}_\text{pt},\text{P}_\text{bb}, \text{P}_\text{mask},\text{P}_\text{txt}\},
    \label{eq:sam-1}
\end{align}
%
where $\mathbf{I}\in \mathds{R}^{H\times W\times 3}$ is the input image and $\text{P}$ denotes a kind of prompt sampled from the set $\mathcal{P}=\{\text{P}_\text{pt},\text{P}_\text{bb}, \text{P}_\text{mask},\text{P}_\text{txt}\}$ including the point, bounding box, mask, and text. 
The output $\mathcal{M}=\{\mathbf{M}_i\in \mathds{R}^{H\times W\times 3}\}_{i=1}^K$ is a set of masks indicating the object we want to segment.
Specifically, we can decompose the $\text{SAM}(\cdot)$ into multiple parts.
An image encoder ($\phi(\cdot)$) is first used to extract the embedding of the input image, and a prompt encoder ($\varphi(\cdot)$) is employed to encode the input prompt and get the prompt embedding.  Then, the image and prompt embeddings are fed to a mask decoder ($\psi(\cdot)$) to generate valid masks. With the above formulations, we can represent \reqref{eq:sam-1} as
%
\begin{align}
    \mathcal{M} = \psi(\phi(\mathbf{I}), \varphi(\text{P})), \text{s.t.}~\text{P}\in\mathcal{P}=\{\text{P}_\text{pt},\text{P}_\text{bb}, \text{P}_\text{mask},\text{P}_\text{txt}\}.
    \label{eq:sam-2}
\end{align}
%
SAM can be transferred to a series of zero-shot tasks like zero-shot single-point valid mask evaluation, zero-shot edge detection, zero-shot object proposal, etc. Although presenting impressive results, some preliminary results show that SAM easily fails under some complex scenes, presenting some robustness issues. As shown in Fig.~\ref{fig:SAM_result_window}, when facing zoom blur, SAM cannot accurately segment the object indicated by the point.

Evaluating the robustness of SAM comprehensively is critical for the downstream applications, which could also trigger a future direction for such an important foundation model. To this end, we propose to study the robustness of SAM in two ways: \textit{First}, we extend existing adversarial attacking methods to SAM, and see whether SAM can present similar segmentation performance under adversarial examples. \textit{Second}, we build a corrupted dataset by adding diverse natural corruptions with different severity levels to clean images. Then, we can evaluate the performance degradation of SAM before and after adding the corruption. In the following two sub-sections, we detail how to conduct attacks and corruption analysis. 

\subsection{Adversarial Attacks against SAM}

We follow existing representative adversarial attacks like projected gradient descent (PGD) \cite{madry2017towards} to generate adversarial examples to mislead SAM. 
Specifically, given an input image $\mathbf{I}$, a prompt $\text{P}$, and the ground truth masks $\mathcal{M}^*$, we aim to generate an adversarial example $\hat{\mathbf{I}}$ that make the SAM generates masks a noticeable difference to $\mathcal{M}^*$. We formulate this objective as
%
\begin{align}
    \hat{\delta} = \arg \text{min}_{\delta} \mathcal{J}(\text{SAM}(\mathbf{I}+\delta, \text{P}), \mathcal{M}^*), \text{s.t.}~\|\delta\|_\text{p} \leq \epsilon
    \label{eq:attack-sam}
\end{align}
%
where $\mathcal{J}(\cdot)$ denotes the objective function to measure the distance between predicted masks and the ground truth ones, $\hat{\delta}$ is the estimated adversarial perturbation whose $L_\text{p}$ norm should be less than the hyper-parameter $\epsilon$, and $\hat{\mathbf{I}} = \mathbf{I} + \hat{\delta}$ is denoted as the generated adversarial example. 
%

%

\subsection{Common Corruptions against SAM}
We add commonly used corruptions to generate corrupted examples to mislead SAM. To be specific, given an input image $\mathbf{I}$, a kind of corruption $c$, and the corruption severity $k$, the generated corrupted image $\hat{\mathbf{I}}$ is 
\begin{align}
    \hat{\mathbf{I}} = \mathcal{C}(\mathbf{I},c,k),
    \label{eq:sam-corruption}
\end{align}
where $\mathcal{C}(\cdot)$ is the function related to the corruption type $c$. For example, in terms of noise corruption, $\mathcal{C}(\cdot)$ is a kind of addition process. 

\section{Benchmark Construction}

\subsection{Adversarial Attacks Selection}
In this work, we employ four gradient-based optimization techniques: FGSM \cite{goodfellow2014explaining}, BIM \cite{kurakin2018adversarial}, PGD \cite{madry2017towards}, and SegPGD \cite{gu2022segpgd} methods to optimize the objective Eq. (\ref{eq:attack-sam}).
Among them, PGD is an iterative attack that can generate strong white-box attacks \cite{guo2021sparta}. 
SegPGD improves the generation of adversarial samples for segmentation models. 
We will try other adversarial attacks in the near future. We focus on $l_{\infty}$-based perturbations, with the identical objective function setting $\mathcal{J}(\cdot)$ as in the original SAM training setup \cite{kirillov2023segany}. This involved supervising mask prediction with a linear combination of Focal Loss \cite{lin2017focal} and Dice Loss \cite{milletari2016v}, with a 20:1 ratio of Focal Loss to Dice Loss. We also conduct an ablation study against adversarial attack using MSE loss \cite{zhang2023attack}.







\begin{figure*}[htbp]
    \centering
    \includegraphics[width=\linewidth]{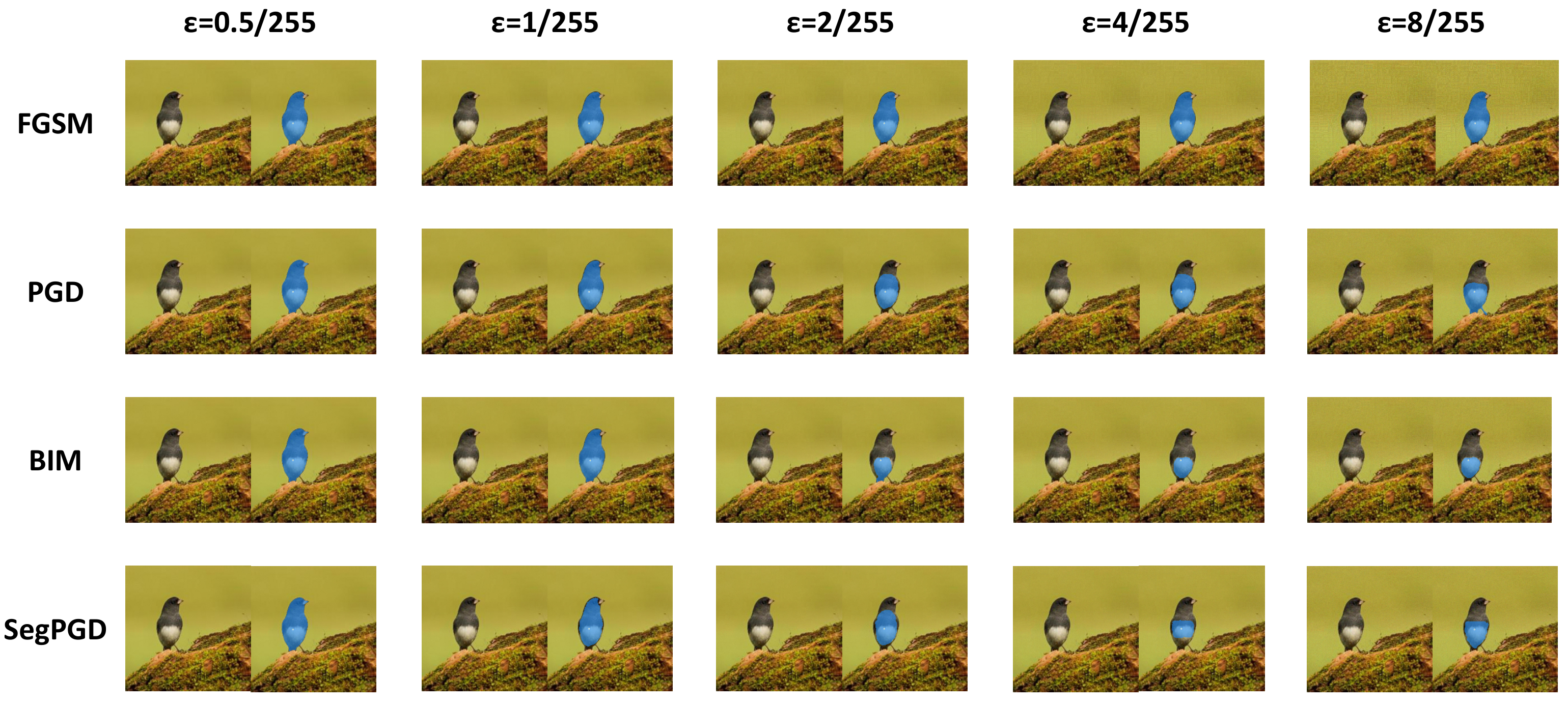}
    \caption{Adversarial attacks examples under 4 kinds of attacks with 5 different severities and corresponding masks predicted by SAM.}
    \label{fig:adv_types}
\end{figure*}

The range of allowed perturbation value $\epsilon$ is set to {0.5/255, 1/255, 2/255, 4/255, 8/255}. The total number of steps is 10 and the step size is set as 1/255 for PGD, BIM, and SegPGD. 
The detailed adversarial attack examples (including corresponding segmentation results by SAM) are shown in Fig.~\ref{fig:adv_types}.

\subsection{Common Corruption Selection}
\begin{figure*}[htbp]
    \centering
    \includegraphics[width=\linewidth]{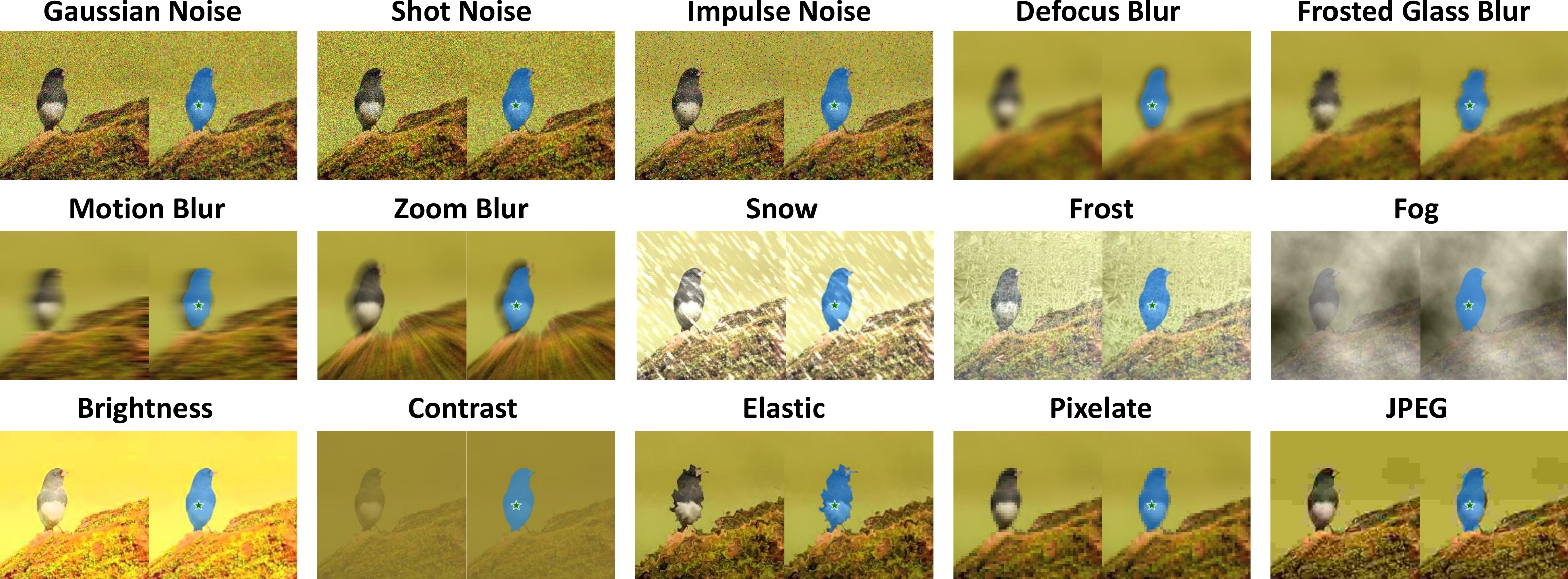}
    \caption{Corruption examples of 15 diverse types and corresponding masks predicted by SAM.}
    \label{fig:corruption_types}
\end{figure*}

We follow the common corruptions selected by ImageNet-C \cite{hendrycks2018benchmarking}. That is, there include 15 diverse corruption categories with 5 severities (from 1 to 5, and bigger is more severe). The corruptions are drawn from four types: noise, blur, weather and digital. The detailed corruption examples (including corresponding segmentation results by SAM) are shown in Fig.~\ref{fig:corruption_types}.

\subsection{Datasets Selection}
We use part of the SA-1B dataset provided by SAM \cite{kirillov2023segany}. To be specific, we use all the images in package ``sa\_000020.tar''. The package contains 11,186 images. We also use the training dataset (200 images) of the KITTI segmentation dataset \cite{Alhaija2018IJCV} as an example to evaluate the performance of SAM in the self-driving dataset since self-driving is a scene that is safety-critical. Thus the robustness of SAM on the self-driving dataset is of great necessity to be evaluated. Furthermore, the objects in KITTI are with ground truth annotations, thus we can evaluate the robustness of SAM on objects rather than on regions that lack semantics as on the SA-1B.

\subsection{Metrics for Robustness Evaluation of SAM}
We adapt the commonly used metrics (\ie, PixelAccuracy (PA) and Intersection over Union (IoU)) in segmentation tasks. To be specific, we use the Mean Pixel Accuracy (mPA) and Mean IoU (mIoU).

\section{Experiment and Analysis}

\subsection{Implementation Details}
\noindent\textbf{Implementation details.} 
We use the ``vit\_h'' version of the SAM model as the target model since it is used as the default version in the demo provided by SAM. All 15 corruption types have 5 different severity levels. In our experiment, we use a point in the image as the prompt to test the segmentation result of SAM. For the sake of illustration, we call the object/region that contains the point to be ``foreground" and the other region to be ``background". All the experiments were run on Ubuntu and an NVIDIA GeForce RTX V100 GPU of 32G RAM.

\noindent\textbf{Evaluation steps.} 
For both the SA-1B and the KITTI, we perform three steps to evaluate the segmentation result of SAM on one image. Here we use the IoU metric as an example.

\ding{182} We randomly select a mask in the annotation (\ie, ground truth label) and randomly select a point in the mask as the input of the SAM model.

\ding{183} By putting the point and image pair into the SAM model, it returns a list of binary masks as the prediction. The masks all contain the input point but with different sizes. 

\ding{184} For a list of binary masks \{$\mathbf{M}_1,\mathbf{M}_2,\cdots,\mathbf{M}_n$\}, we calculate the corresponding IoU for each mask and achieve a list of results \{$\mathbf{IoU}_1,\mathbf{IoU}_2,\cdots,\mathbf{IoU}_n$\}. To evaluate the upper segmentation performance of SAM, we select the maximum segmentation evaluation result. That is, 
\begin{equation}
\mathbf{IoU}_{max} = max \{\mathbf{IoU}_1,\mathbf{IoU}_2,\cdots,\mathbf{IoU}_n \},
\label{eq:Cal_Metric_Per_Image}
\end{equation}
where $n$ is the total number of predicted binary masks for one image. Intuitively, for the entire dataset which contains $N$
images, the evaluation result is 
\begin{equation}
\mathbf{IoU} = \frac{1}{N} \sum_{i=1}^{N} \mathbf{IoU}^{i}_{max},
\label{eq:Cal_Metric_Images}
\end{equation}
where $\mathbf{IoU}^{i}_{max}$ represents the IoU result of the $i$th image.

\subsection{Results of Adversarial Attacks against SAM}

\begin{figure}[tbp]
    \centering
    \includegraphics[width=\linewidth]{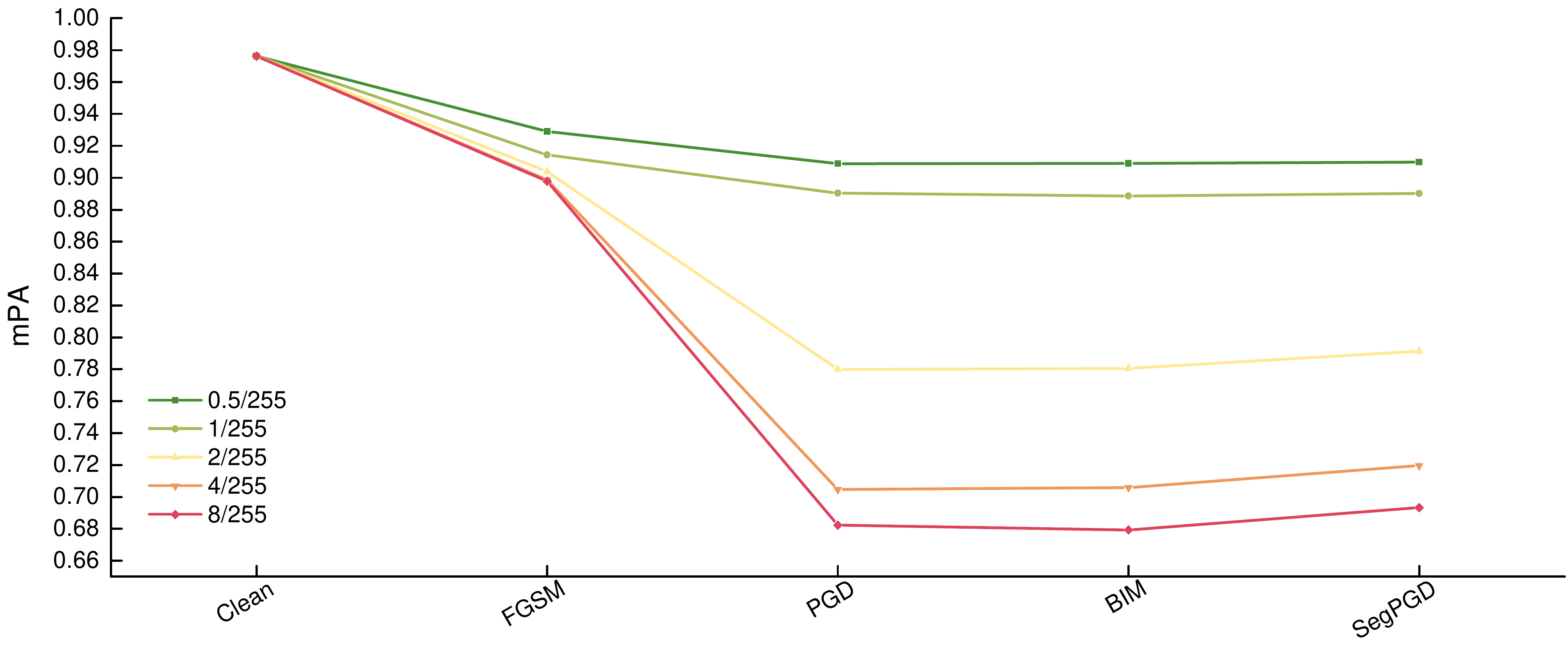}
    \caption{The mPA values of SAM on SA-1B under 4 adversarial attacks and 5 severities.}
    \label{fig:adv_SAM_mPA}
\end{figure}
\begin{figure}[tbp]
    \centering
    \includegraphics[width=\linewidth]{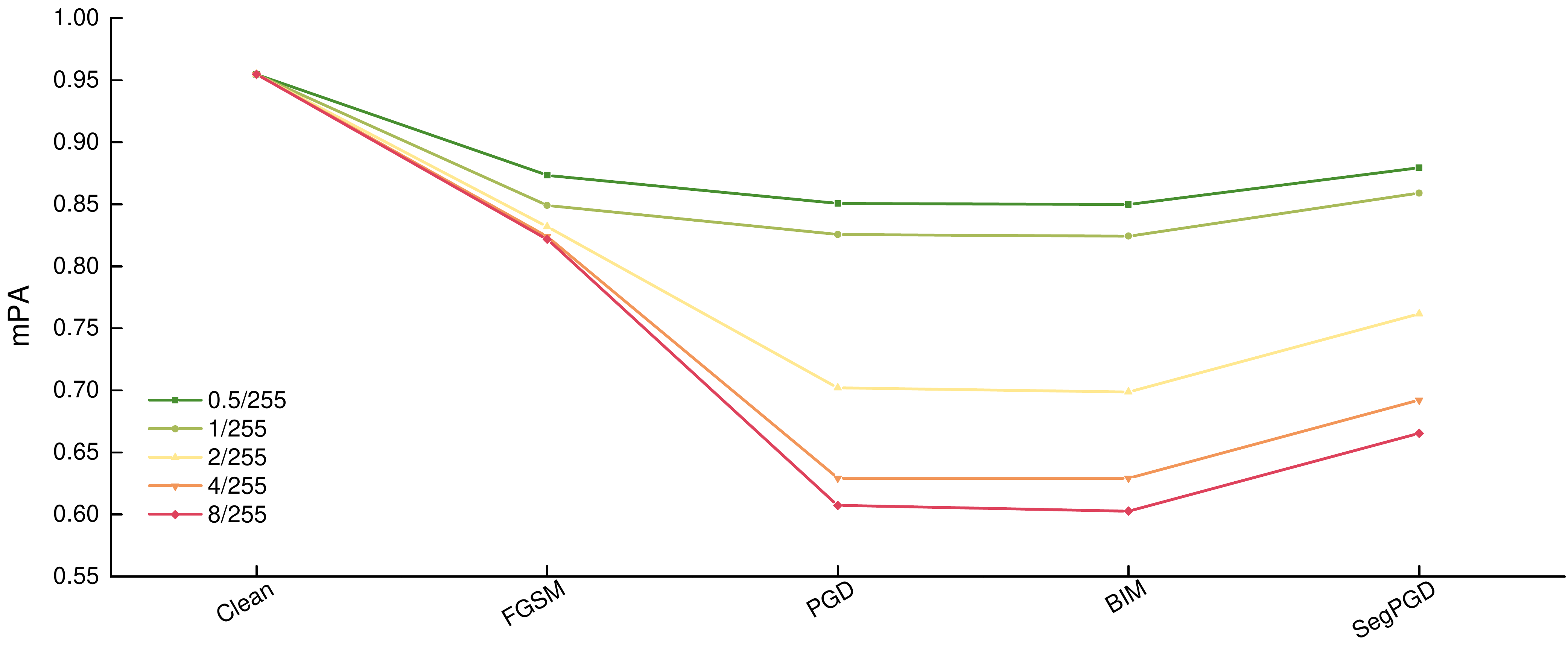}
    \caption{The mIoU values of SAM on SA-1B under 4 adversarial attacks and 5 severities.}
    \label{fig:adv_SAM_mIoU}
\end{figure}

\begin{figure}[tbp]
    \centering
    \includegraphics[width=\linewidth]{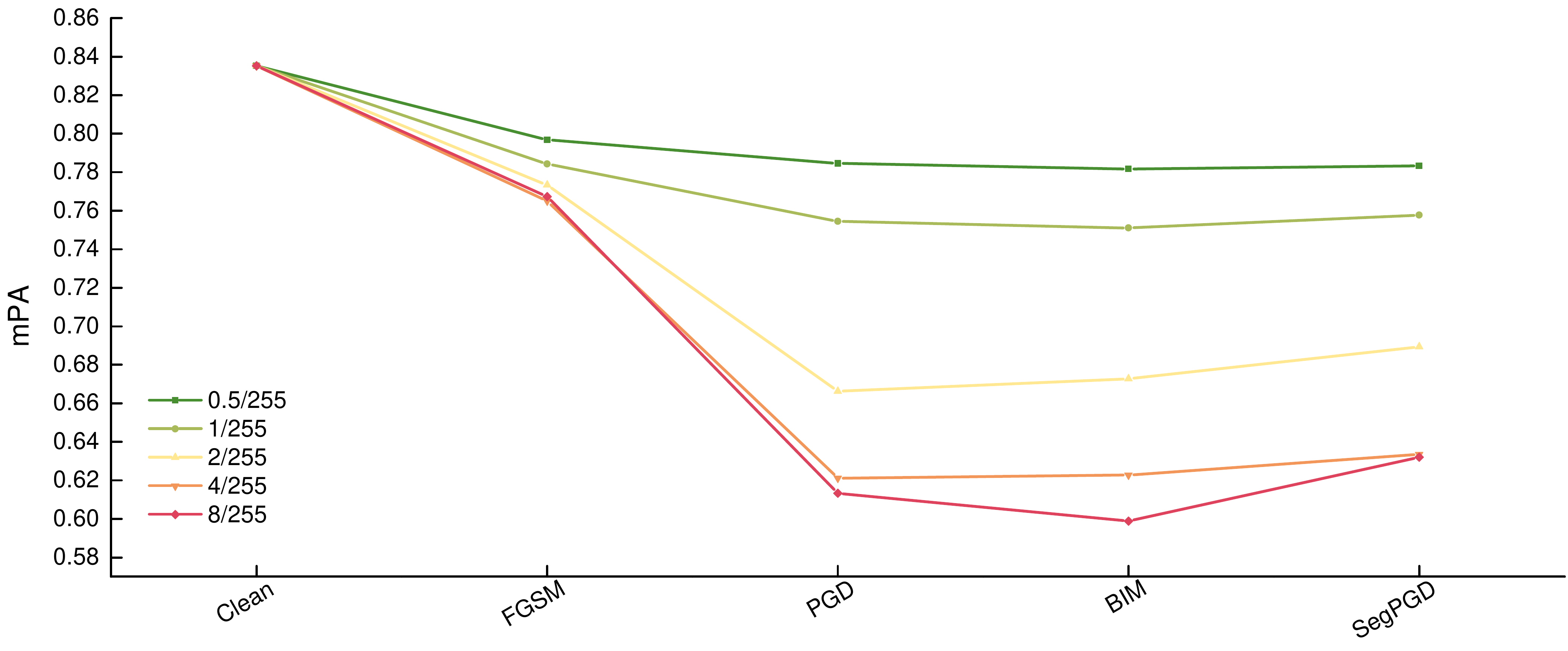}
    \caption{The mPA values of SAM on KITTI under 4 adversarial attacks and 5 severities.}
    \label{fig:adv_KITTI_mPA}
\end{figure}
\begin{figure}[tbp]
    \centering
    \includegraphics[width=\linewidth]{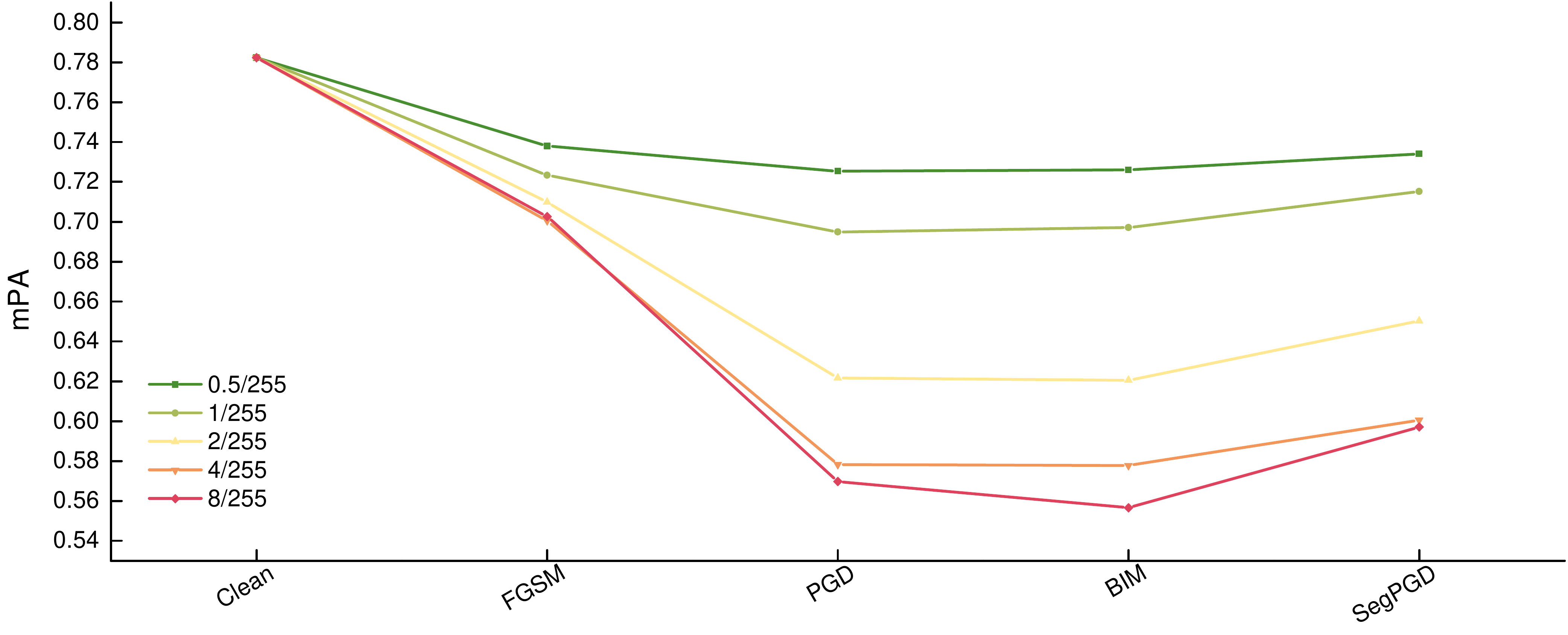}
    \caption{The mIoU values of SAM on KITTI under 4 adversarial attacks and 5 severities.}
    \label{fig:adv_KITTI_mIoU}
\end{figure}

\begin{figure}[tbp]
    \centering
    \includegraphics[width=\linewidth]{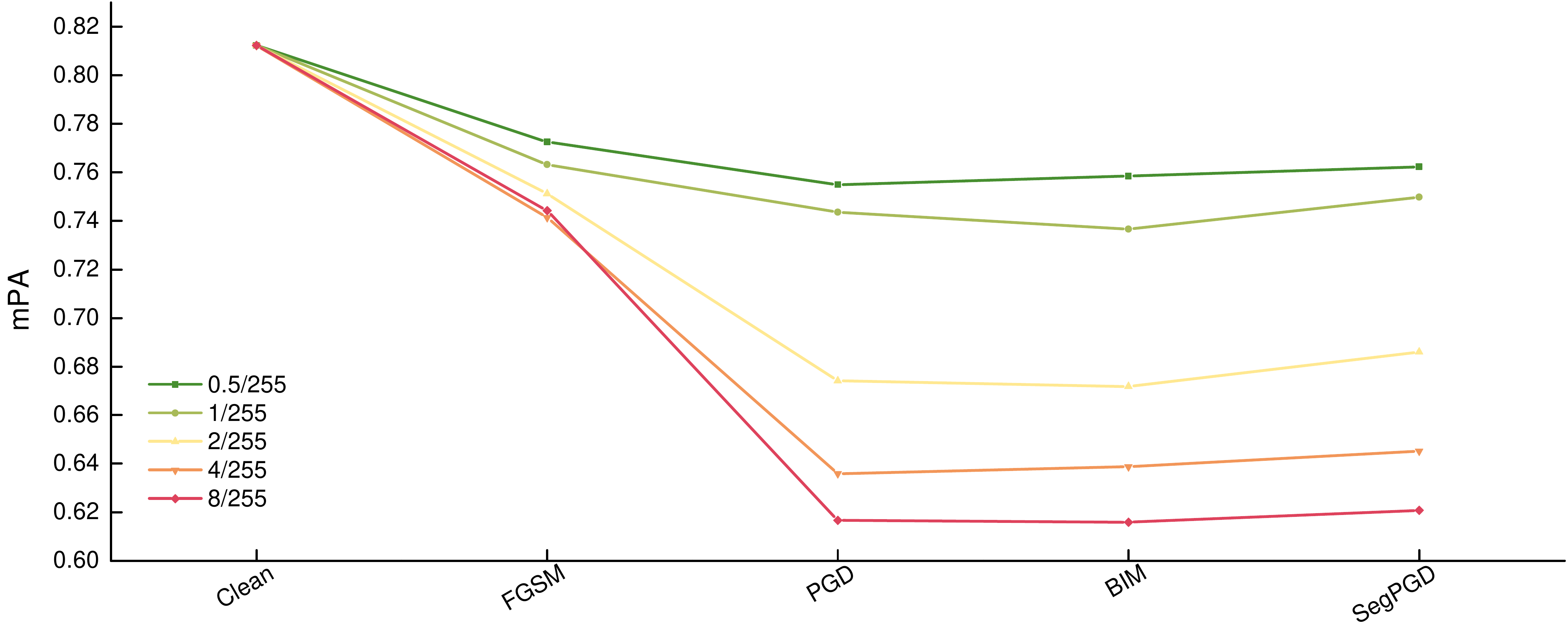}
    \caption{The mPA values of SAM on big objects of KITTI under 4 adversarial attacks and 5 severities.}
    \label{fig:adv_KITTI_big_mPA}
\end{figure}
\begin{figure}[tbp]
    \centering
    \includegraphics[width=\linewidth]{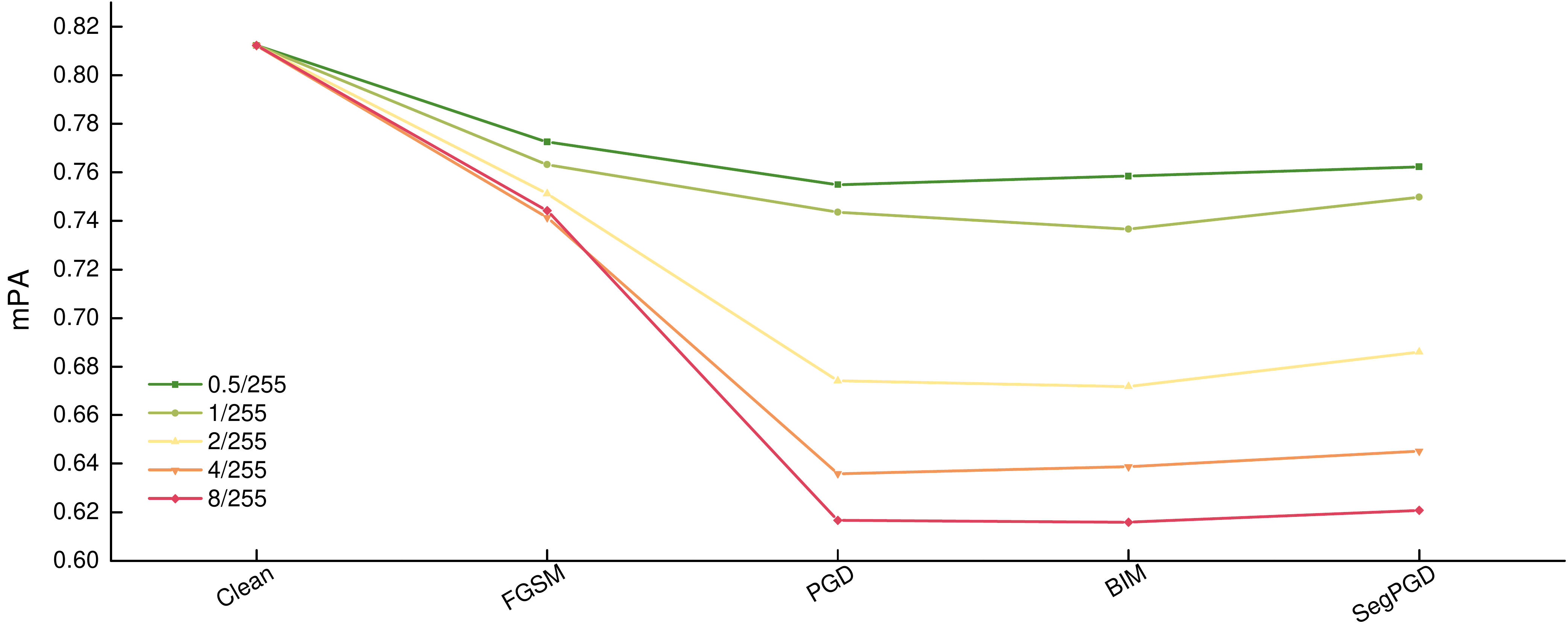}
    \caption{The mIoU values of SAM on big objects of KITTI under 4 adversarial attacks and 5 severities.}
    \label{fig:adv_KITTI_big_mIoU}
\end{figure}

\begin{figure}[tbp]
    \centering
    \includegraphics[width=\linewidth]{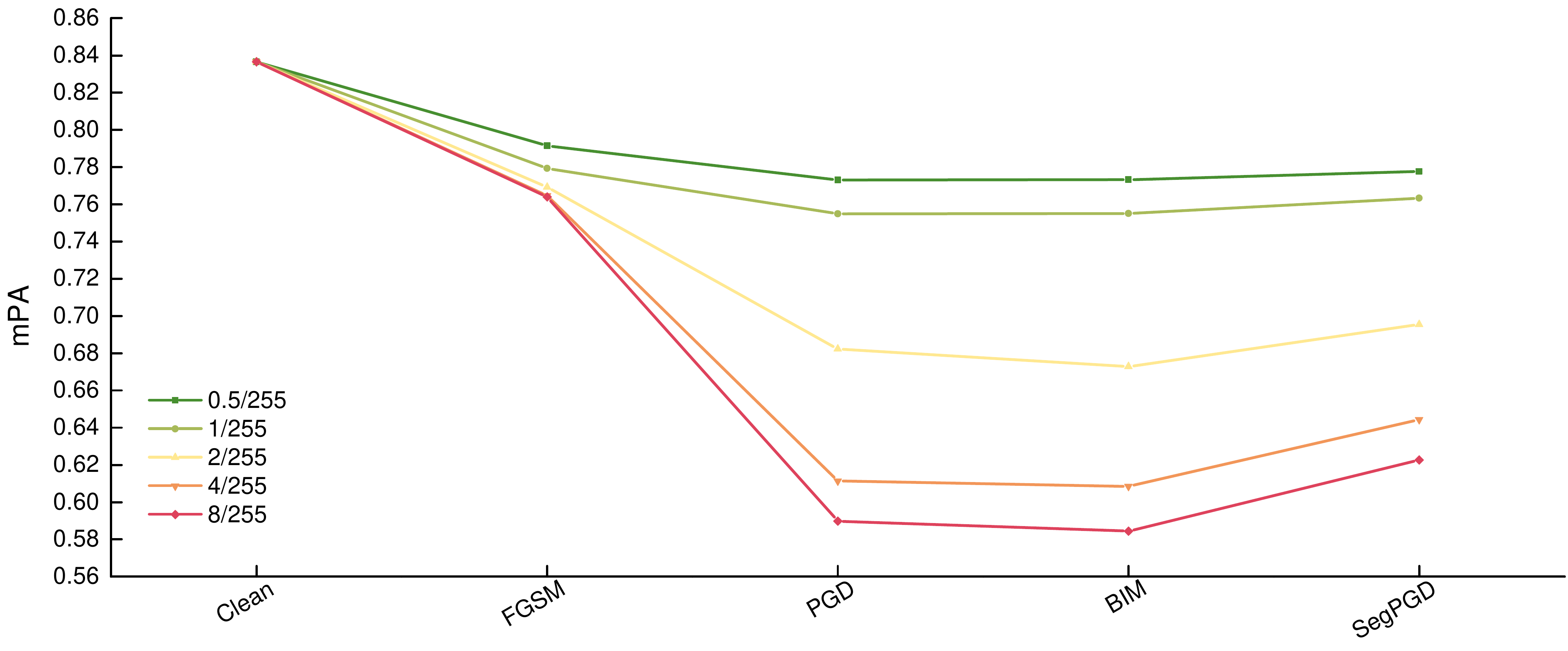}
    \caption{The mPA values of SAM on small objects of KITTI under 4 adversarial attacks and 5 severities.}
    \label{fig:adv_KITTI_small_mPA}
\end{figure}
\begin{figure}[tbp]
    \centering
    \includegraphics[width=\linewidth]{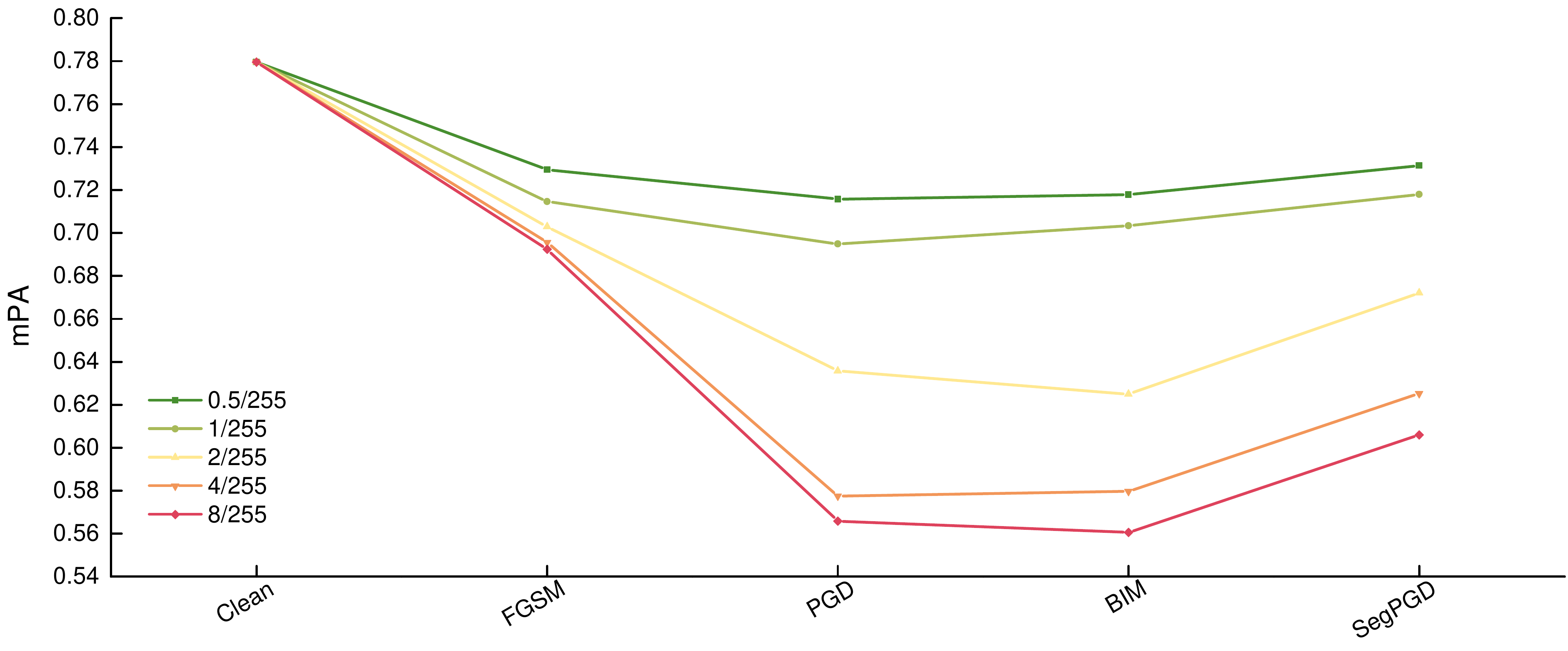}
    \caption{The mIoU values of SAM on small objects of KITTI under 4 adversarial attacks and 5 severities.}
    \label{fig:adv_KITTI_small_mIoU}
\end{figure}

\textbf{Results on SA-1B.}
Fig. \ref{fig:adv_SAM_mPA} and Fig. \ref{fig:adv_SAM_mIoU} depict the mPA and mIoU performance of SA-1B under four different types of adversarial attacks. Table \ref{table:sam_adv} presents the details for mPA, mIoU, PA background, PA foreground, IoU background, and IoU foreground under four types of adversarial attacks on the SA-1B.

As a whole, adversarial attacks generally result in a decrease in both mPA and mIoU for SAM. In the clean (non-adversarial) case, the mPA value is 0.976, indicating a high level of pixel accuracy, while the mIoU value is 0.954, suggesting a high degree of overlap between the predicted and ground truth segmentation masks. The application of adversarial attacks results in a reduction of 0.047 to 0.297 in mPA and 0.075 to 0.352 in mIoU. Moreover, as the $\epsilon$ value increases (indicating a larger perturbation), the degradation in performance becomes more pronounced. Furthermore, the decrease in mPA and mIoU primarily corresponds to foreground segmentation, which indicates that adversarial attacks undermine the semantic information in the image.

Regarding the different types of adversarial attacks, Fig. \ref{fig:adv_SAM_mPA} and Fig. \ref{fig:adv_SAM_mIoU} indicate that the performance drop varies across attack methods, suggesting that different attacks have varying levels of effectiveness on SAM. BIM and PGD consistently exhibit the strongest attack performance. For instance, when $\epsilon$ is set to 8/255, the mPA performance of BIM and PGD degrades to approximately 0.680 compared to the mPA of clean data (i.e., 0.976). In contrast, FGSM could only reduce the mPA value to 0.897, and the mPA under SegPGD attack could be reduced to 0.693, which is in proximity to those of BIM and PGD. Similarly, comparable observations can be made concerning the metric MIoU.

\noindent\textbf{Results on KITTI.}
Fig.~\ref{fig:adv_KITTI_mPA} and Fig.~\ref{fig:adv_KITTI_mIoU} illustrate the mPA and mIoU results of KITTI when subjected to four distinct forms of adversarial attacks. Table \ref{table:kitti_adv} displays the details of mPA, mIoU, PA background, PA foreground, IoU background, and IoU foreground under four types of adversarial attacks on the KITTI.

Overall, adversarial attacks also lead to a decrease in both mPA and mIoU for KITTI. With lower mPA and mIoU values (0.835 and 0.782) in a clean situation compared to the SA-1B, adversarial attacks decrease the mPA by 0.038 to 0.236 and the mIoU by 0.044 to 0.225.

As the $\epsilon$ value increases, the mPA and mIoU values tend to decrease for most attacks. This suggests that the segmentation performance deteriorates as the perturbations become stronger. 
However, when the intensity of the attack reaches a certain threshold, the magnitude of the decrease becomes less noticeable.

Furthermore, different attacks exhibit varying performances in the KITTI. As depicted in Table \ref{table:kitti_adv}, both BIM and PGD consistently exhibit the strongest attack performance. In particular, BIM demonstrates the highest degradation of 0.236 in the mPA metric. Additionally, the mIoU metric experiences a decline from 0.782 to 0.667.

\noindent\textbf{Results on big objects of KITTI.}
Figures \ref{fig:adv_KITTI_big_mPA} and \ref{fig:adv_KITTI_big_mIoU} depict the mPA and mIoU outcomes of big objects of KITTI when exposed to four different types of adversarial attacks. Table \ref{table:kitti_big_adv} displays the detailed results of mPA, mIoU, PA background, PA foreground, IoU background, and IoU foreground under four types of adversarial attacks on relatively big objects in KITTI.

In general, the application of adversarial attacks on relatively big objects in KITTI results in a reduction of both mPA and mIoU. When compared to the KITTI random mask dataset, the clean situation yields lower mPA and mIoU values of 0.812 and 0.776, respectively. As a consequence of adversarial attacks, the mPA decreases by 0.039 to 0.196, while the mIoU decreases by 0.041 to 0.209.

As the $\epsilon$ value is raised, there is a general trend of decreasing mPA and mIoU values across most attacks. Nevertheless, as the attack intensity increases to a certain extent, the magnitude of the decrease becomes subtle.

Additionally, experiments on relatively big objects in KITTI shows varying performance across different attacks. As shown in Table \ref{table:kitti_big_adv}, both BIM and PGD consistently demonstrate stronger attack performance. Specifically, BIM exhibits the highest degradation of 0.196 in the mPA metric. Moreover, the mIoU metric undergoes a decrease, falling from 0.776 to 0.567.

\noindent\textbf{Results on small objects of KITTI.} 
The results of small objects of KITTI under four different types of adversarial attacks are illustrated in Figures \ref{fig:adv_KITTI_small_mPA} and \ref{fig:adv_KITTI_small_mIoU}, showcasing the mPA and mIoU metrics respectively. Table \ref{table:kitti_small_adv} shows the detailed results of mPA, mIoU, PA background, PA foreground, IoU background, and IoU foreground under four types of adversarial attacks on relatively small objects of KITTI.

Generally, when adversarial attacks are applied to small objects of KITTI, both mPA and mIoU decrease. In the absence of adversarial attacks, the mPA value is 0.836 and the mIoU value is 0.779, which is slightly higher (with an increase of 0.024 and 0.003, respectively) compared to the performance of big objects of KITTI. Adversarial attacks lead to a reduction in the mPA ranging from 0.045 to 0.252 and a decrease in mIoU  ranging from 0.048 to 0.218. Moreover, as the $\epsilon$ value increases, there is a consistent trend of decreasing mPA and mIoU values observed across the various attacks.

Moreover, experiments on small objects of KITTI exhibit diverse performance levels when subjected to different attacks. Table \ref{table:kitti_small_adv} illustrates the consistent and stronger attack performance of both BIM and PGD. Notably, BIM showcases the highest degradation with a decrease of 0.252 in the mPA metric. Furthermore, there is a decrease in the mIoU metric, dropping from 0.779 to 0.560.

\subsection{Ablation Study of Adversarial Attacks on MSE Loss}

\begin{figure}[tbp]
    \centering
    \includegraphics[width=\linewidth]{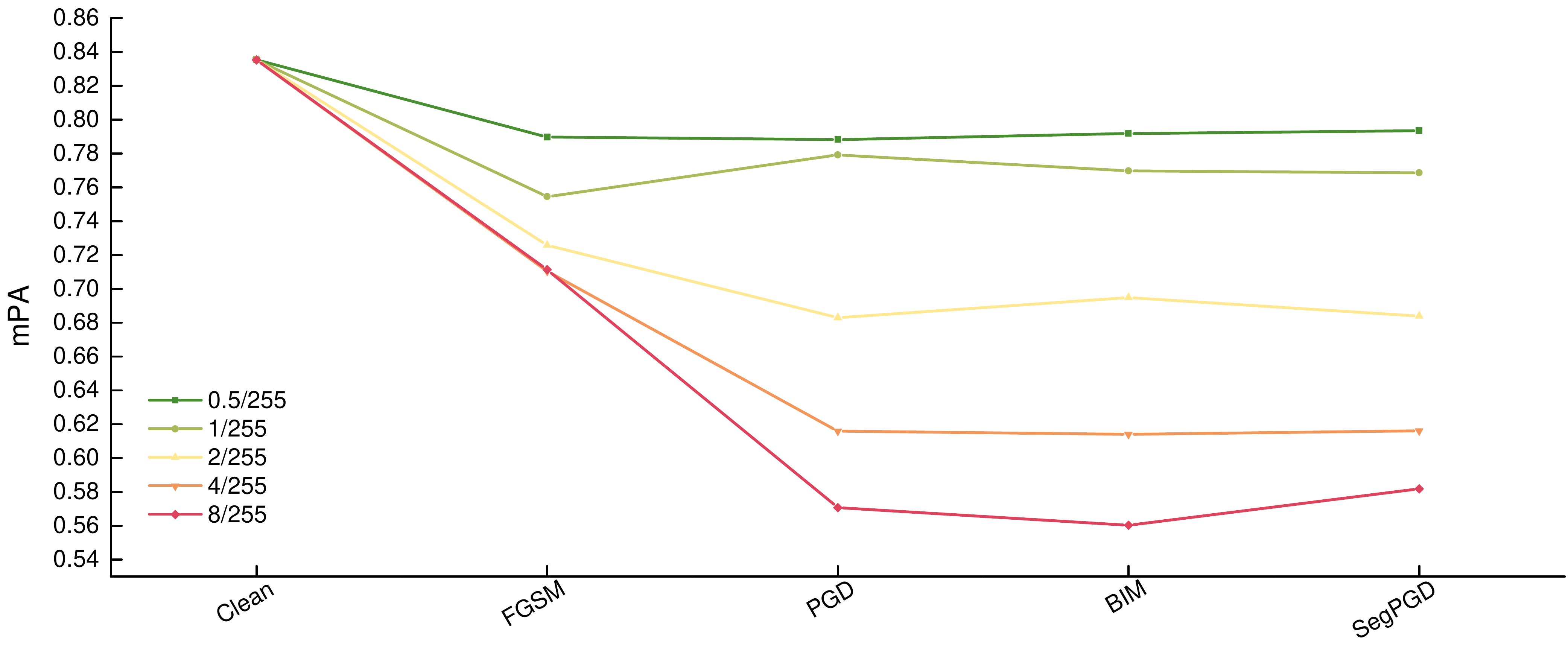}
    \caption{The mPA values of SAM on KITTI under 4 adversarial attacks and 5 severities using MSE loss.}
    \label{fig:adv_mse_KITTI_mPA}
\end{figure}
\begin{figure}[tbp]
    \centering
    \includegraphics[width=\linewidth]{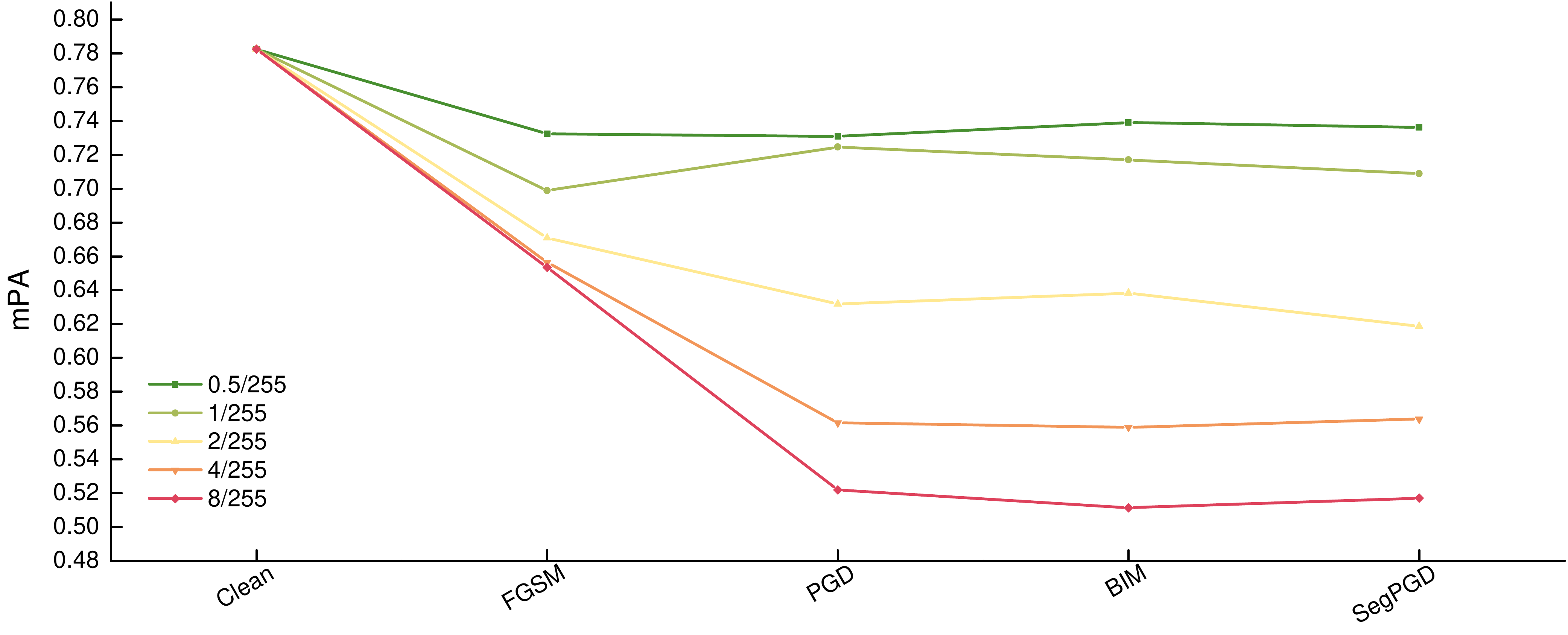}
    \caption{The mIoU values of SAM on KITTI under 4 adversarial attacks and 5 severities using MSE loss.}
    \label{fig:adv_mse_KITTI_mIoU}
\end{figure}
\begin{figure}[tbp]
    \centering
    \includegraphics[width=\linewidth]{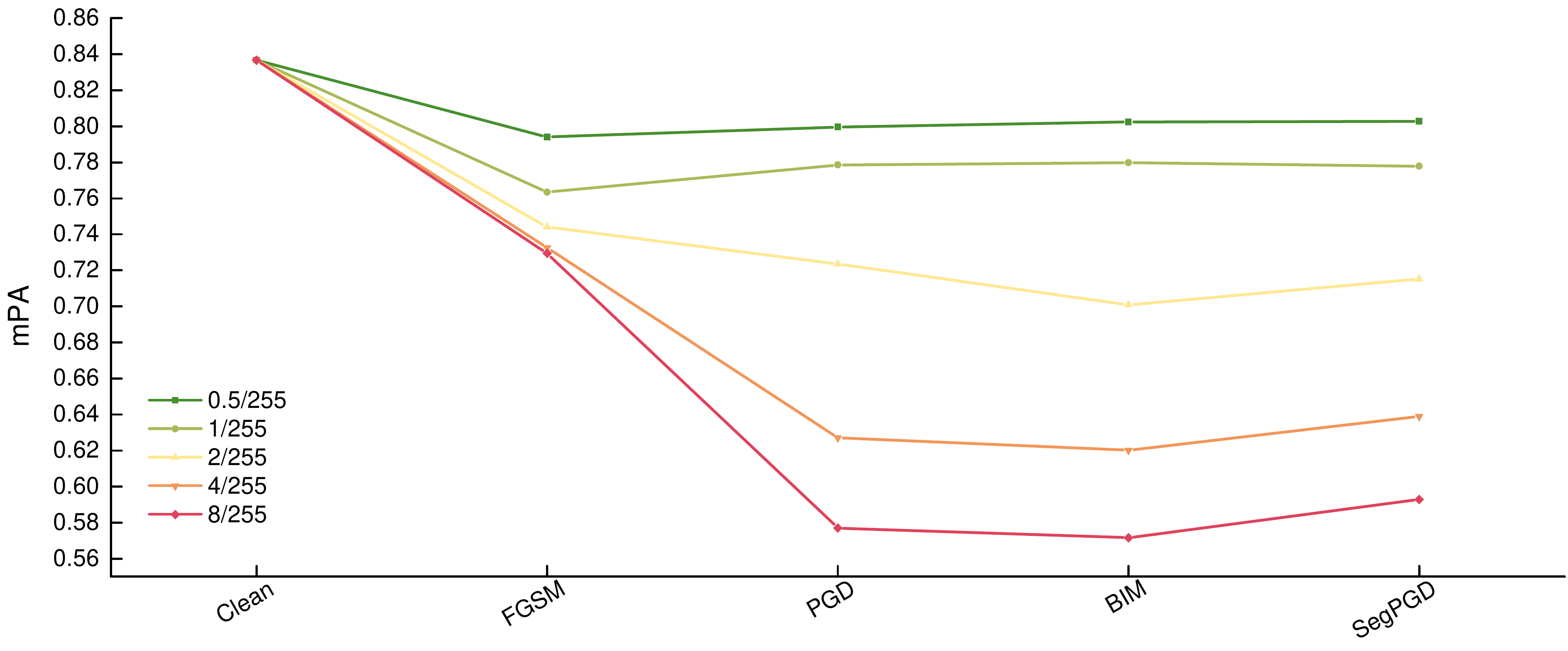}
    \caption{The mPA values of SAM on big objects of KITTI under 4 adversarial attacks and 5 severities using MSE loss.}
    \label{fig:adv_mse_KITTI_big_mPA}
\end{figure}
\begin{figure}[tbp]
    \centering
    \includegraphics[width=\linewidth]{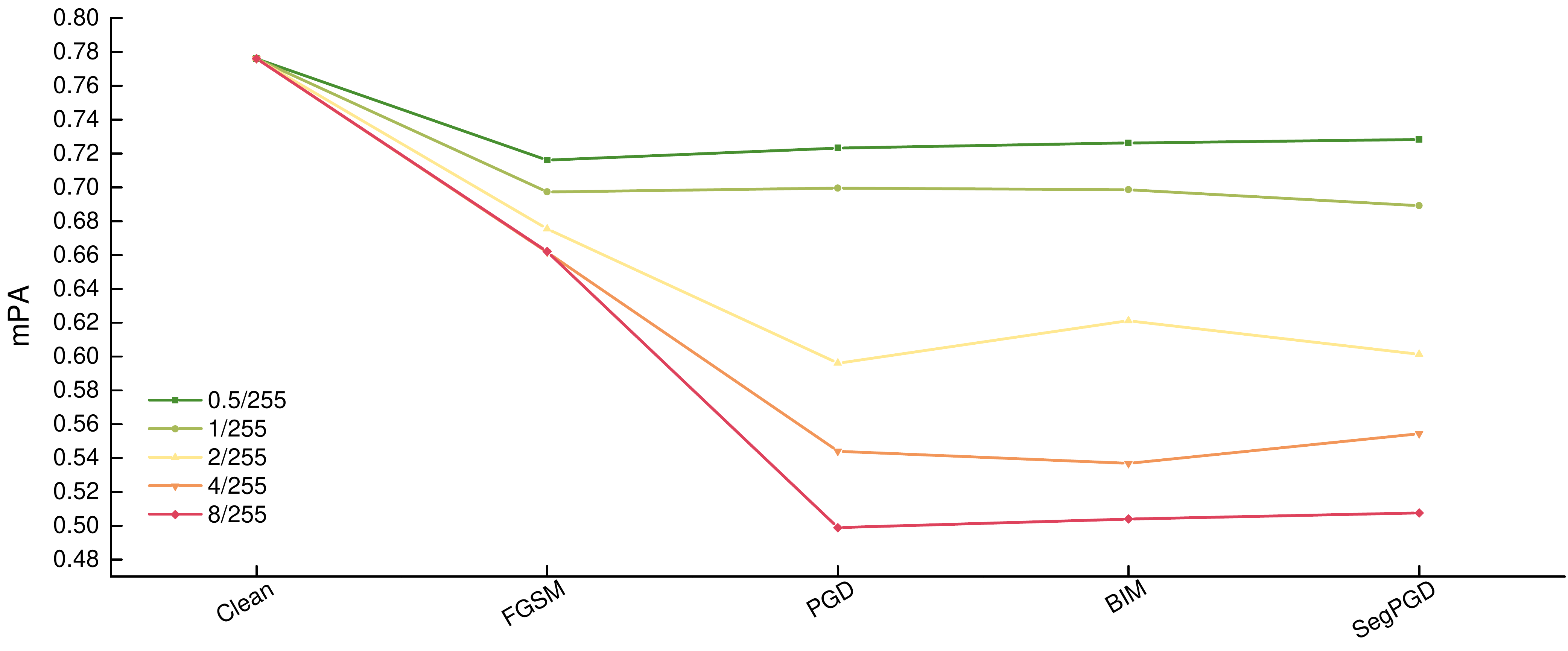}
    \caption{The mIoU values of SAM on big objects of KITTI under 4 adversarial attacks and 5 severities using MSE loss.}
    \label{fig:adv_mse_KITTI_big_mIoU}
\end{figure}

\begin{figure}[tbp]
    \centering
    \includegraphics[width=\linewidth]{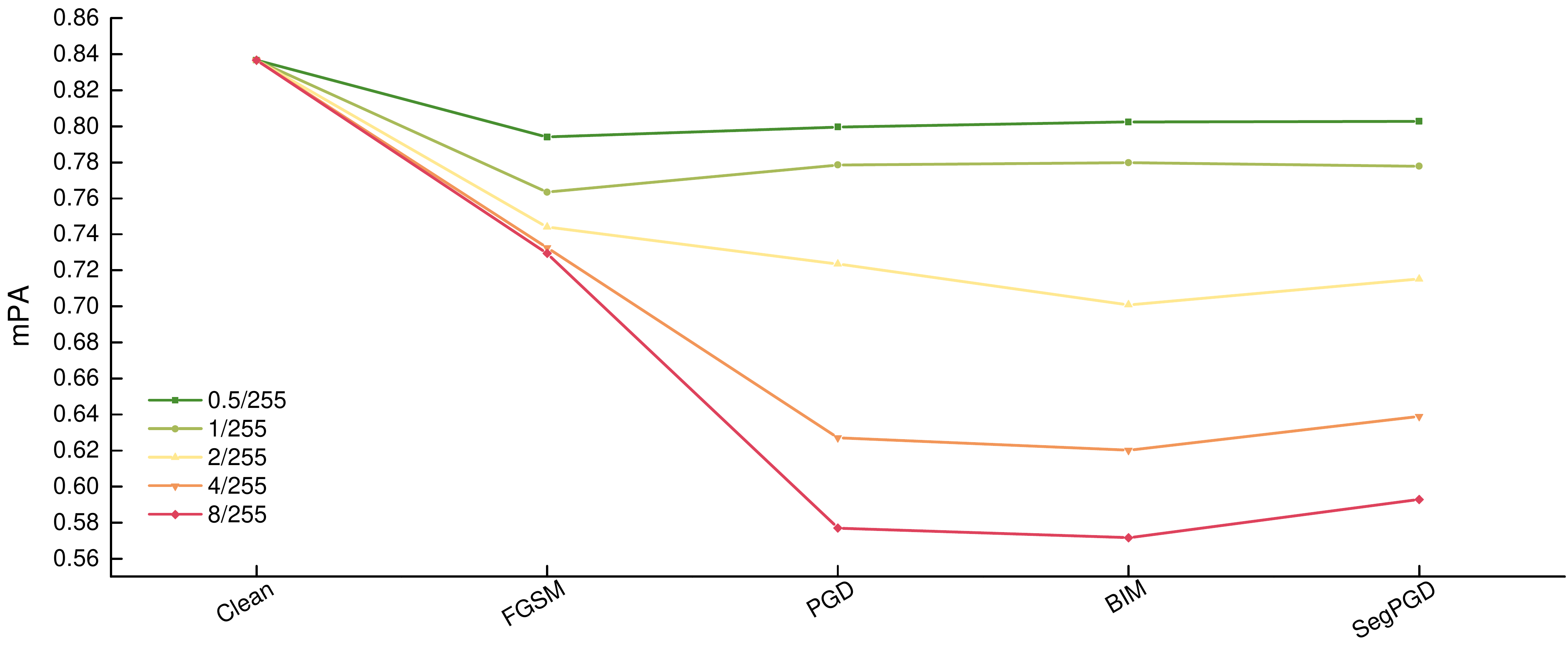}
    \caption{The mPA values of SAM on small objects of KITTI under 4 adversarial attacks and 5 severities using MSE loss.}
    \label{fig:adv_mse_KITTI_small_mPA}
\end{figure}
\begin{figure}[tbp]
    \centering
    \includegraphics[width=\linewidth]{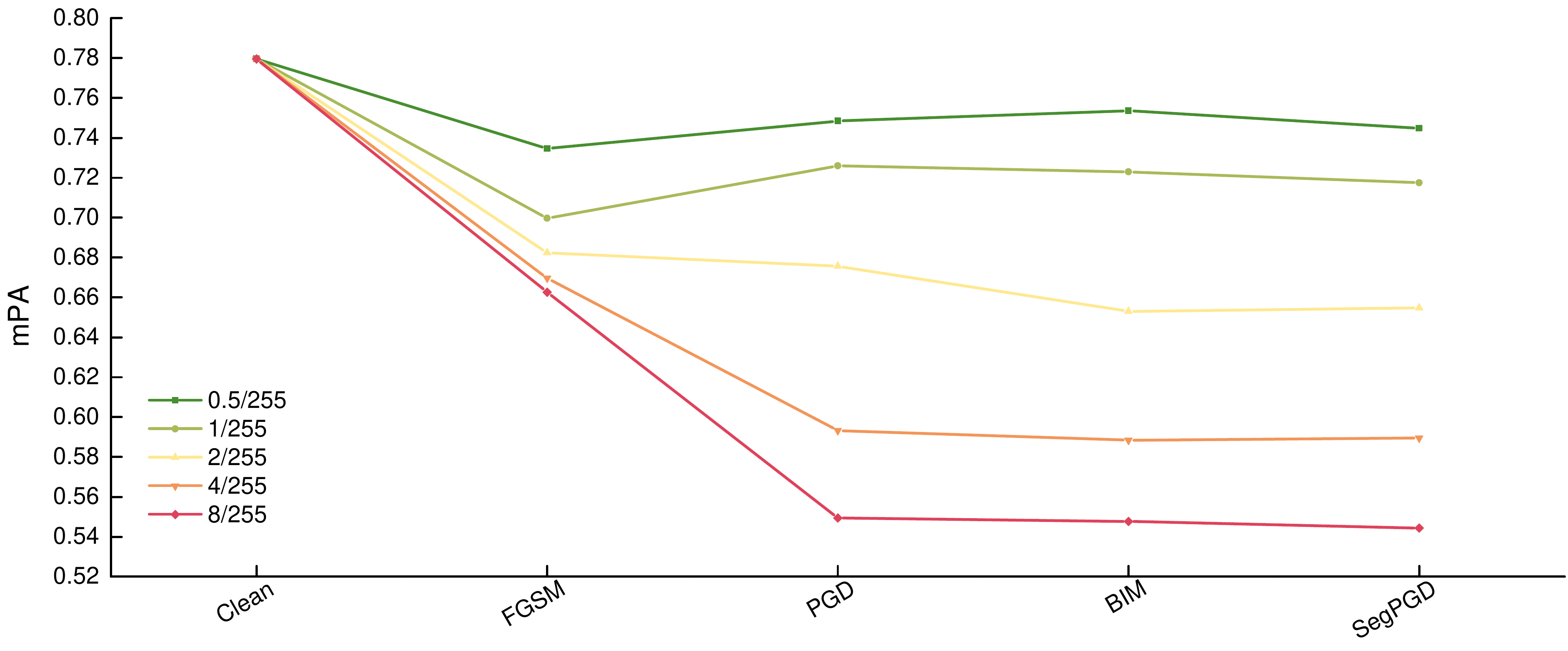}
    \caption{The mIoU values of SAM on small objects of KITTI under 4 adversarial attacks and 5 severities using MSE loss.}
    \label{fig:adv_mse_KITTI_small_mIoU}
\end{figure}

In this section, we set $\mathcal{J}(\cdot)$ in Eq. \ref{eq:attack-sam} as mean square error (MSE) loss, as presented in \cite{zhang2023attack}, to generate adversarial samples. We apply this loss to the KITTI dataset and obtain comparable results to those reported in prior research.

The mPA and mIoU results of KITTI, big objects of KITTI, and small objects of KITTI are presented in Fig. \ref{fig:adv_mse_KITTI_mPA}, \ref{fig:adv_mse_KITTI_mIoU}, \ref{fig:adv_mse_KITTI_big_mPA}, \ref{fig:adv_mse_KITTI_big_mIoU}, \ref{fig:adv_mse_KITTI_small_mPA},
\ref{fig:adv_mse_KITTI_small_mIoU}. More detailed results are in Table \ref{table:mse_kitti_adv}, \ref{table:mse_kitti_big_adv} and \ref{table:mse_kitti_small_adv}.

\subsection{Results of Common Corruptions against SAM}

\begin{figure}[tbp]
    \centering
    \includegraphics[width=\linewidth]{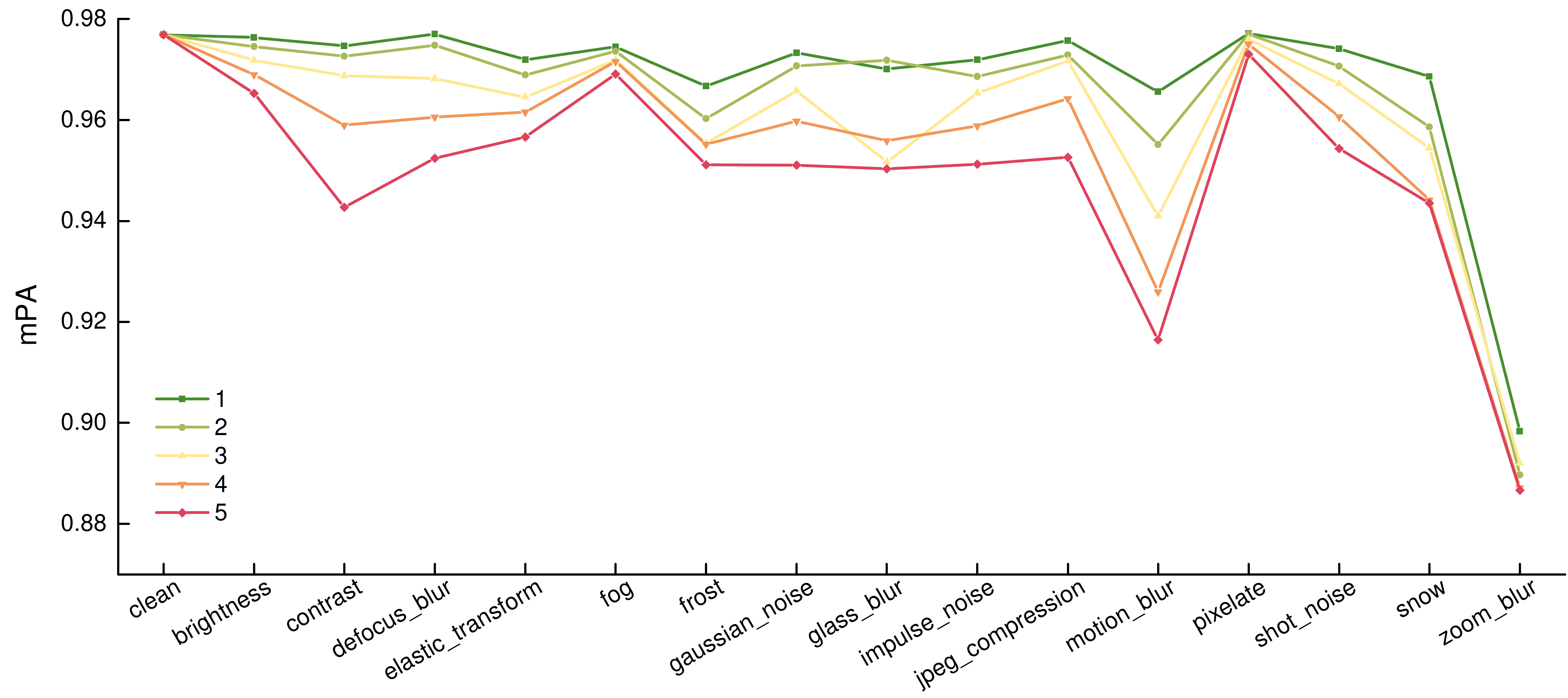}
    \caption{The mPA values of SAM on SA-1B under 15 corruptions and 5 severities.}
    \label{fig:SAM_mPA}
\end{figure}
\begin{figure}[tbp]
    \centering
    \includegraphics[width=\linewidth]{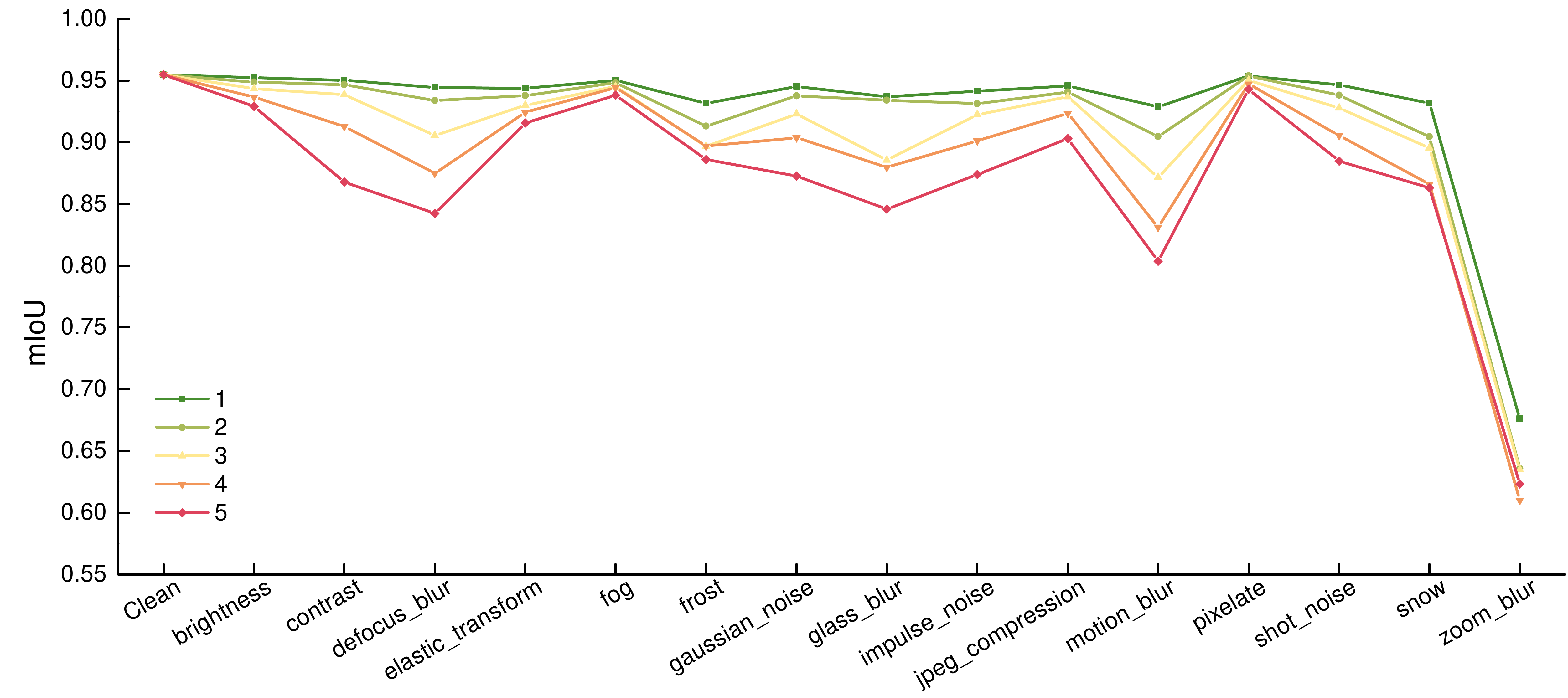}
    \caption{The mIoU values of SAM on SA-1B under 15 corruptions and 5 severities.}
    \label{fig:SAM_mIoU}
\end{figure}
\begin{figure}[tbp]
    \centering
    \includegraphics[width=\linewidth]{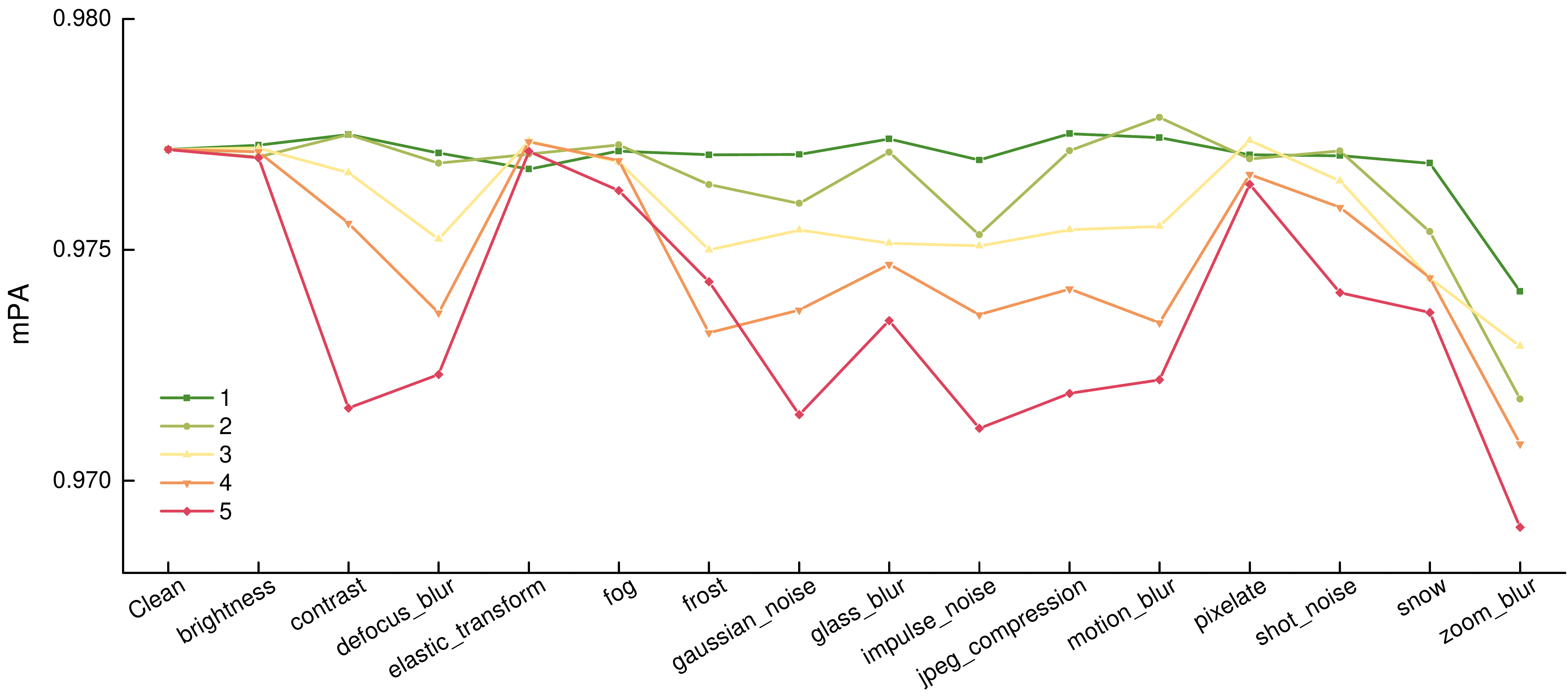}
    \caption{The mPA values of SAM on KITTI under 15 corruptions and 5 severities.}
    \label{fig:KITTI_mPA}
\end{figure}
\begin{figure}[tbp]
    \centering
    \includegraphics[width=\linewidth]{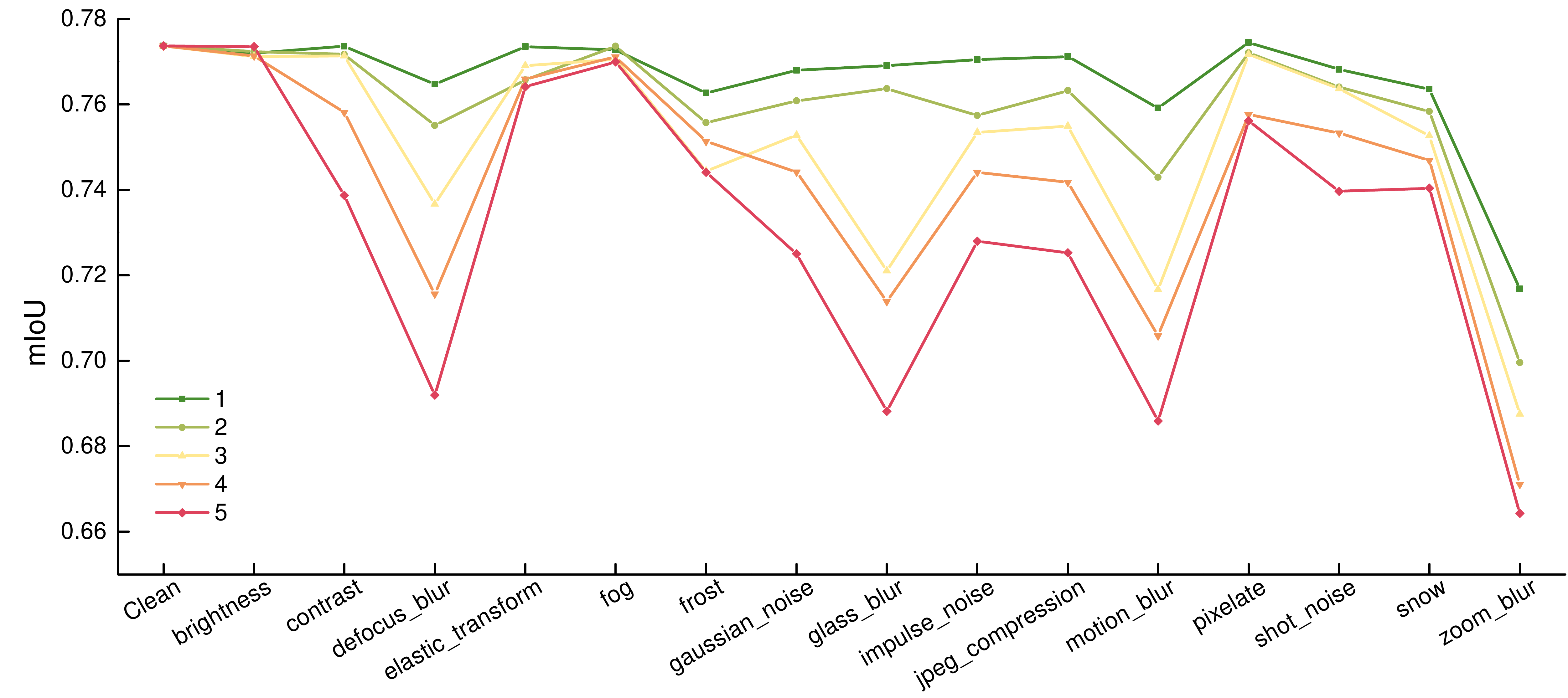}
    \caption{The mIoU values of SAM on KITTI under 15 corruptions and 5 severities.}
    \label{fig:KITTI_mIoU}
\end{figure}
\begin{figure}[tbp]
    \centering
    \includegraphics[width=\linewidth]{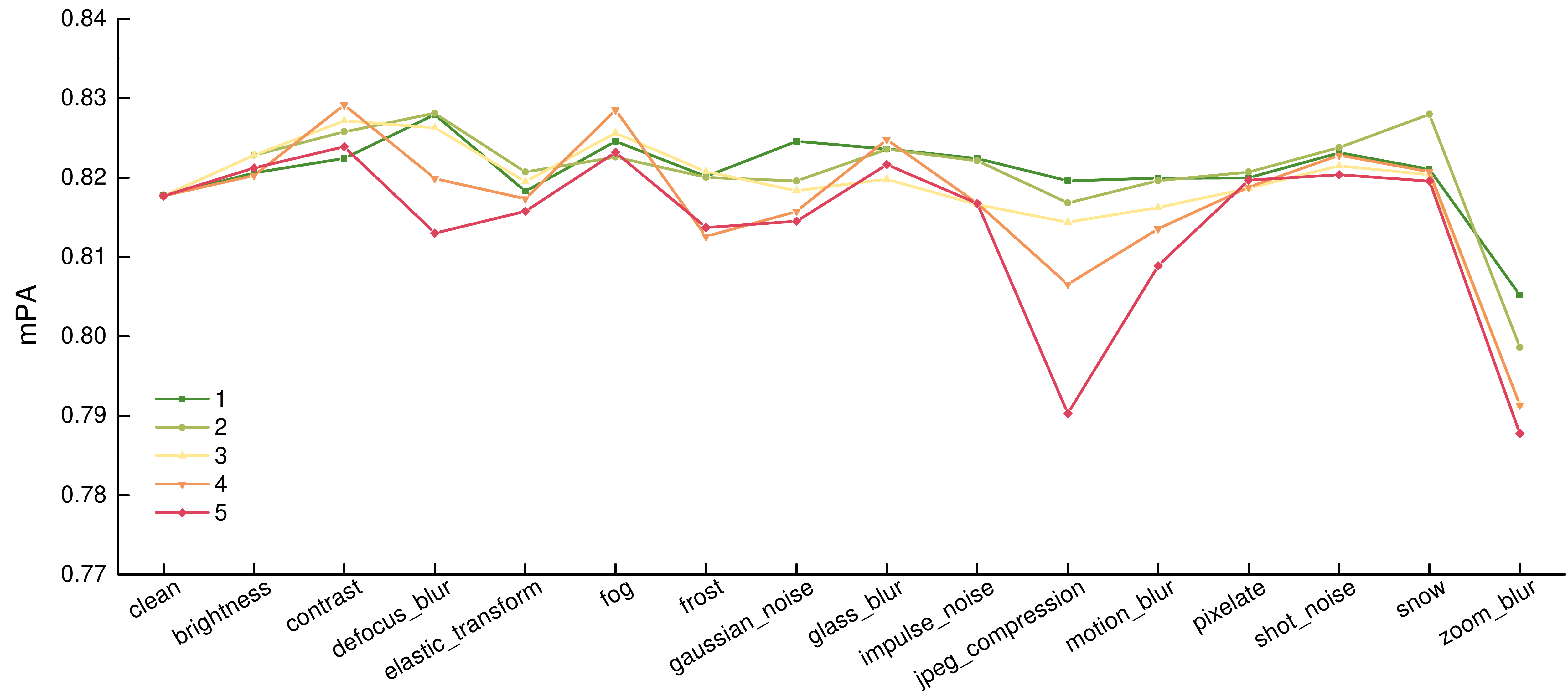}
    \caption{The mPA values of SAM on big objects of KITTI under 15 corruptions and 5 severities.}
    \label{fig:KITTI_big_mPA}
\end{figure}
\begin{figure}[tbp]
    \centering
    \includegraphics[width=\linewidth]{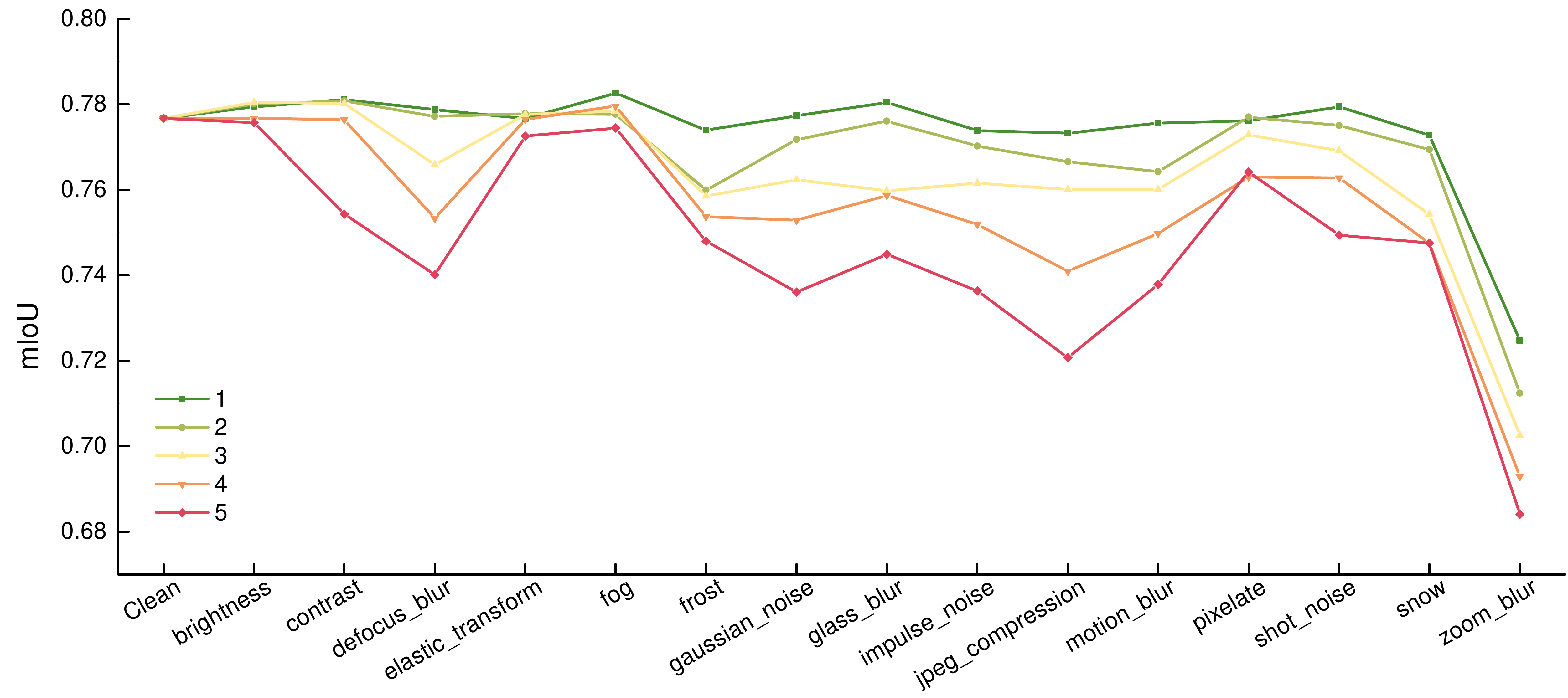}
    \caption{The mIoU values of SAM on big objects of KITTI under 15 corruptions and 5 severities.}
    \label{fig:KITTI_big_mIoU}
\end{figure}

\begin{figure}[tbp]
    \centering
    \includegraphics[width=\linewidth]{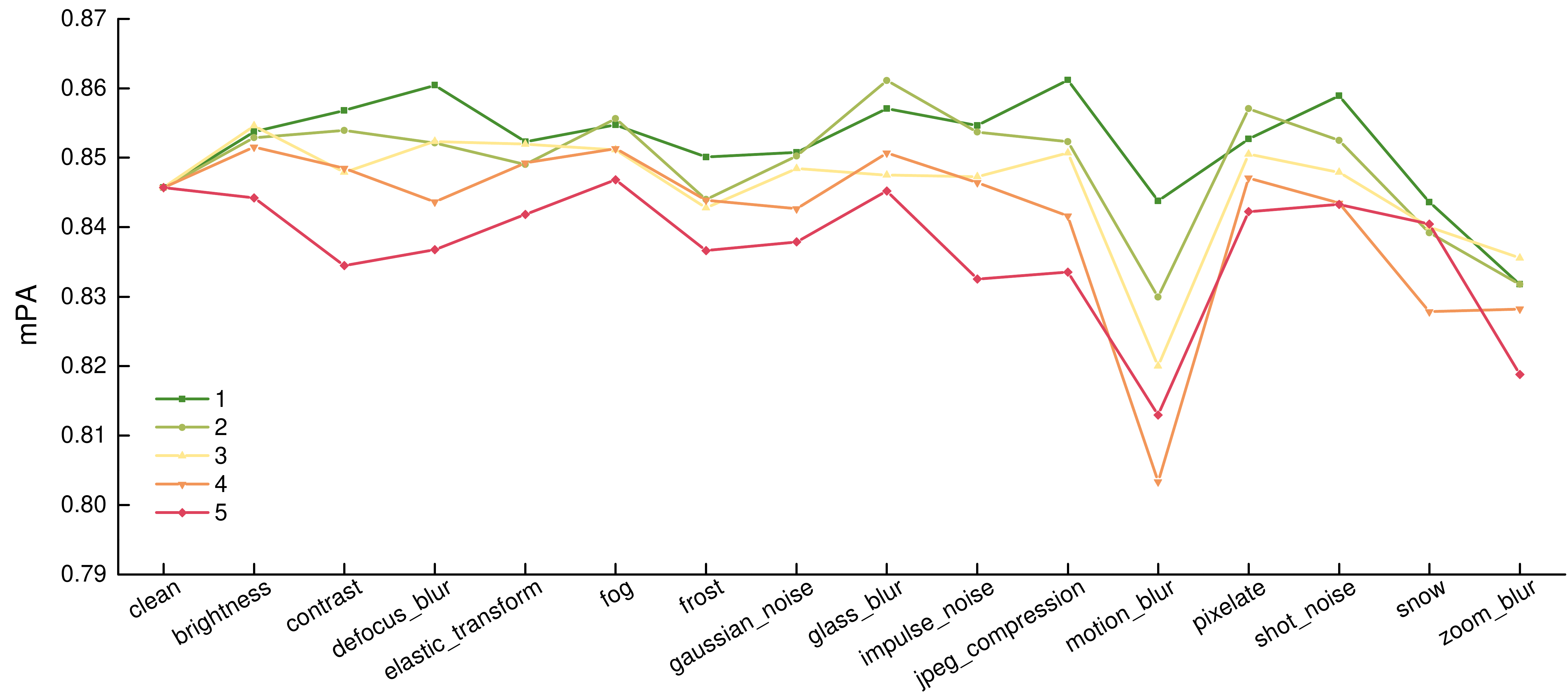}
    \caption{The mPA values of SAM on small objects of KITTI under 15 corruptions and 5 severities.}
    \label{fig:KITTI_small_mPA}
\end{figure}
\begin{figure}[tbp]
    \centering
    \includegraphics[width=\linewidth]{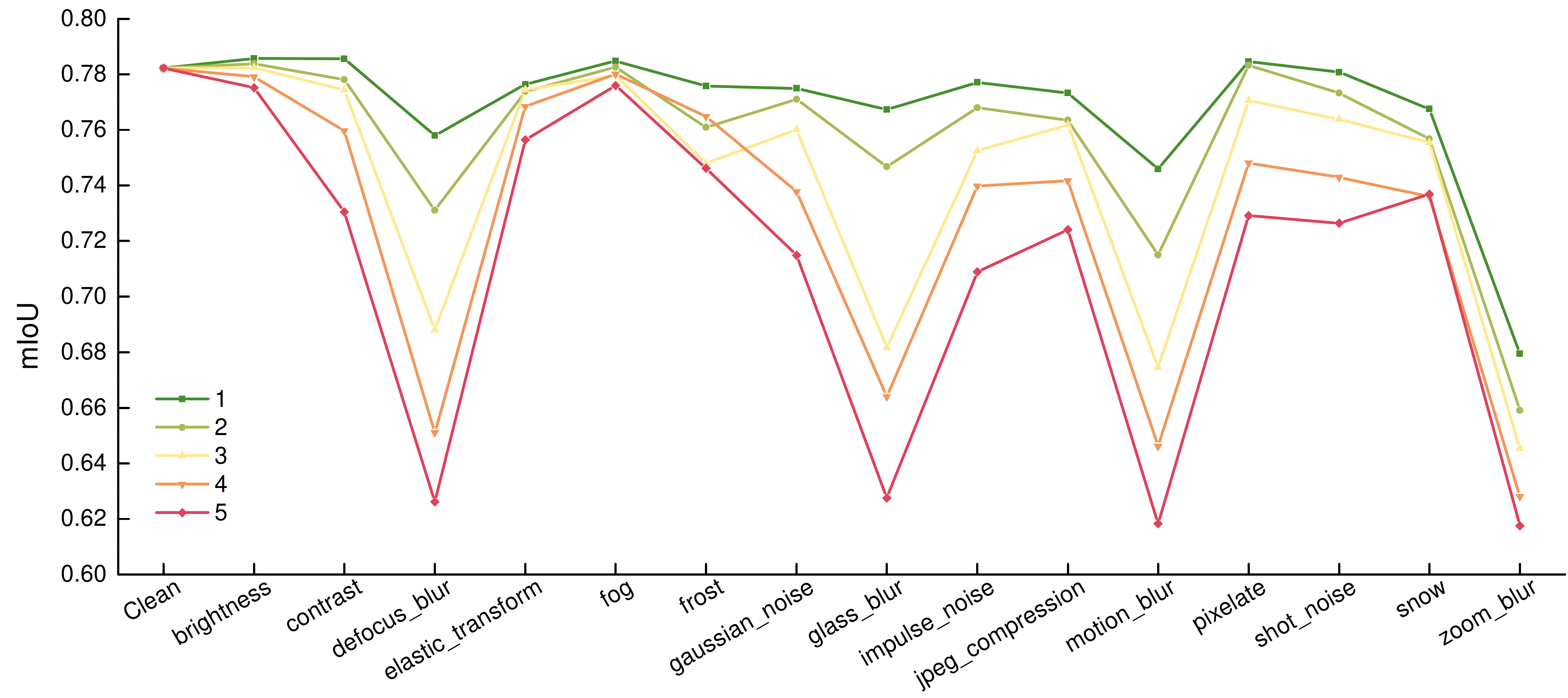}
    \caption{The mIoU values of SAM on small objects of KITTI under 15 corruptions and 5 severities.}
    \label{fig:KITTI_small_mIoU}
\end{figure}

\textbf{Results on SA-1B.} As shown in Fig.~\ref{fig:SAM_mPA}, on the whole, the SAM has excellent robustness against corruption. We can find that the decline in each metric and each corruption is minor, even under the biggest corruption severity. To be specific, in terms of the mPA metric, under the biggest corruption severity, in most of the corruptions, the decline is less than 5\% and the decrease in zoom blur is the largest (still less than 10\%). As shown in Fig.~\ref{fig:SAM_mIoU}, in terms of the mIoU metric, under the biggest corruption severity, in most of the corruptions, the decline is less than 12\% and the decrease in zoom blur is the largest. From the figure, among the corruptions, we can find that blur-related corruptions can affect the performance of SAM the most. In contrast, among both mPA and mIoU metrics, pixelate and fog have a very minor influence on the performance of SAM. The detailed mPA and mIoU metric values are shown in Table~\ref{tab:sam_corruption_mean}.

mPA and mIoU just show the average performance of SAM. To show the performance in detail, we also provide PA and IoU results of SAM on ``foreground'' and ``background'' regions in Table~\ref{tab:sam_corruption_pa_detail} and Table~\ref{tab:sam_corruption_iou_detail}. We can find that the SAM has excellent robustness against corruption in background regions. In most cases, the PA and IoU values are almost no decline. In contrast, the performance of SAM in the foreground region is a bit worse. Please note that no matter PA or IoU metric, the SAM has a huge loss of performance on motion blur and zoom blur, which promotes us to focus on blur-related corruptions when aim to improve the robustness of SAM.

\noindent\textbf{Results on KITTI.}
As shown in Fig.~\ref{fig:KITTI_mPA}, we can find that SAM achieves extremely good performance on mPA under all the corruption types and corruption severities. The decline is less than 2\% in all the situations. As shown in Fig.~\ref{fig:KITTI_mIoU}, the robustness is also good on the mIoU metric (in most cases, the decline are less than 5\%.) exclude blur-related corruptions (\ie, motion blur, zoom blur, defocus blur and frosted glass blur). The detailed mPA and mIoU metric values are shown in Table~\ref{tab:kitti_corruption_mean}.

mPA and mIoU just show the average performance of SAM. To show the performance in detail, we also provide PA and IoU results of SAM on ``foreground'' and ``background'' regions in Table~\ref{tab:kitti_corruption_pa_detail} and Table~\ref{tab:kitti_corruption_iou_detail}. In terms of background region, we can find that the performance of SAM on PA and IoU is extremely high under all the corruptions types. With respect to the foreground region, the performance of SAM on the PA metric is not high (less than 70\%) but robust. In contrast, the performance on the IoU metric is low (less than 60\%) and not as robust as on the PA metric. The performance declined significantly on blur-related corruptions (more than 15\%) and some noise-related corruptions (\ie, Gaussian noise and impulse noise) and JPEG corruption (more than 5\%).

\noindent\textbf{Results on big objects of KITTI.} To evaluate the performance and robustness of SAM on objects of different scales, we conduct experiments on relatively big objects in KITTI scenes. To be specific, we select the point from the masks of the top 50\% size in an image of KITTI. As shown in Fig.~\ref{fig:KITTI_big_mPA} and Fig.~\ref{fig:KITTI_big_mIoU}, the robustness of SAM is extremely well across almost all the corruption types on mPA and mIoU metric. The performance is only a bit lower on mIoU against zoom blur than others. The detailed mPA and mIoU metric values are shown in Table~\ref{tab:kitti_corruption_big_mask_mean}.

With respect to detailed values on the foreground object and background region in Table~\ref{tab:kitti_corruption_big_mask_pa_detail} and \ref{tab:kitti_corruption_big_mask_iou_detail}, we can find that the performance on background region is significantly good. All the results are higher than 90\% across all the corruption types on both PA and IoU metrics. In terms of foreground objects, the performance of SAM is not high but the robustness is good on the PA metric. Most of the decline is less than 5\%. In contrast, the robustness of SAM on the IoU metric is a bit worse than on the PA metric (most are less than 10\%), especially on zoom blur corruption (more than 15\% decline).

\noindent\textbf{Results on small objects of KITTI.} To evaluate the performance and robustness of SAM on objects of different scales, we conduct experiments on relatively small objects in KITTI scenes. To be specific, we select the point from the masks of the bottom 50\% size in an image of KITTI. As shown in Fig.~\ref{fig:KITTI_small_mPA} and Fig.~\ref{fig:KITTI_small_mIoU}, the robustness of SAM is extremely well across almost all the corruption types on mPA and mIoU metric. The performance is only a bit lower on mIoU against blur-related corruptions (\ie, defocus blur, frosted glass blur, motion blur, zoom blur). The detailed mPA and mIoU metric values are shown in Table~\ref{tab:kitti_corruption_small_mask_mean}.

With respect to detailed values on the foreground object and background region in Table~\ref{tab:kitti_corruption_small_mask_pa_detail} and \ref{tab:kitti_corruption_small_mask_iou_detail}, we can find that the performance on background region is significantly good. All the results are higher than 90\% across all the corruption types on both PA and IoU metrics. In terms of foreground objects, the performance of SAM is not high but the robustness is good. Most of the decline is less than 7\%. In contrast, the robustness of SAM on the IoU metric is a bit worse than on PA metric (most are less than 15\%). However, the robustness is bad on blur-related corruptions since the decline is more than 30\%.

To summarize, we find that the robustness of SAM against corruption is generally good and the performance is stable on arbitrary scale regions/objects. However, SAM is not good at handling blur-related corruption, which should be carefully focused in future research.

\subsection{Visualization Results}

\begin{figure*}[htbp]
    \centering
    \includegraphics[width=\linewidth]{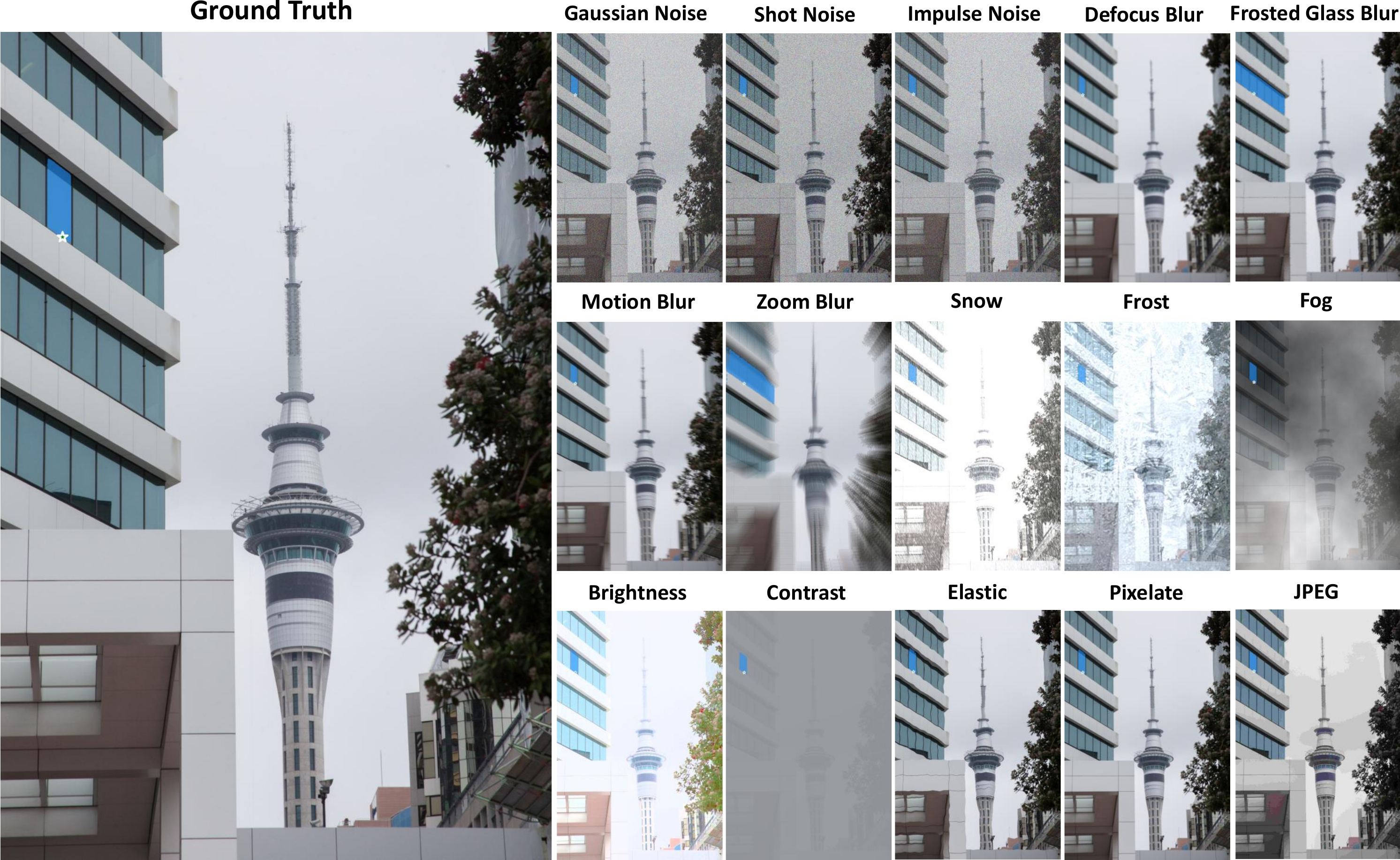}
    \caption{The visualization results on an image of SA-1B. This figure shows the result on a window (a region of regular edges). The segmentation results against corruption are excellent excluding zoom blur and glass blur.}
    \label{fig:SAM_result_window}
\end{figure*}

\begin{figure*}[htbp]
    \centering
    \includegraphics[width=\linewidth]{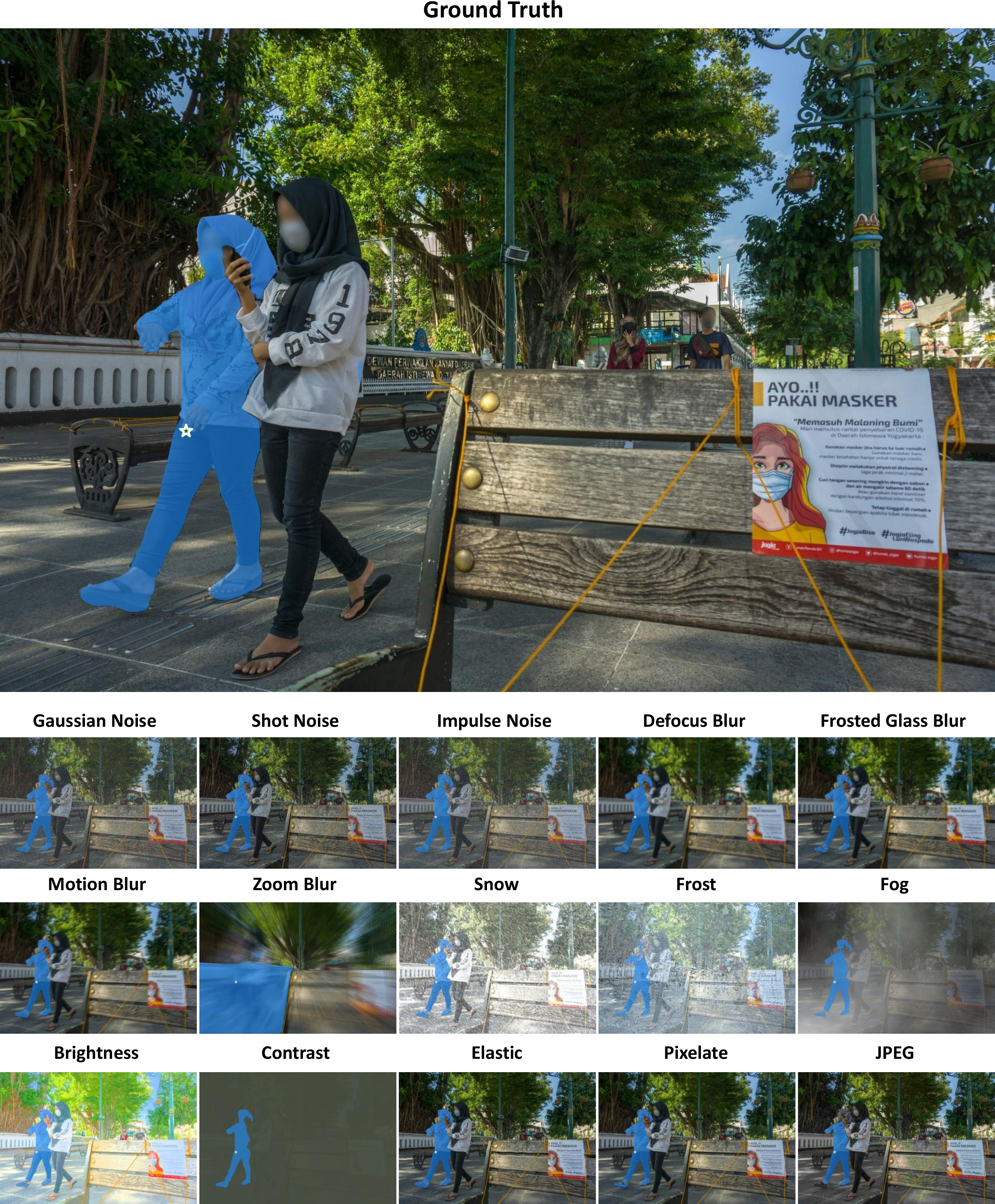}
    \caption{The visualization results on an image of SA-1B. This figure shows the result on a person (a region of irregular edges). The segmentation results against corruption are excellent excluding zoom blur.}
    \label{fig:SAM_result_person}
\end{figure*}

\noindent\textbf{Results on SA-1B against corruption.} As shown in Fig.~\ref{fig:SAM_result_window}, the SAM model is required to predict the segmentation result with the given point (\ie, the star in the image). We can find that SAM achieves excellent results in segmenting the window (a region with regular edges) under most of the corruption with the strongest severity. It only fails when facing glass blur and zoom blur. As shown in Fig.~\ref{fig:SAM_result_person}, we can find that SAM achieves excellent results in segmenting the person (a region with irregular edges) under most of the corruption with the strongest severity. It only fails when facing zoom blur. We can find that zoom blur actually changes the image significantly, thus the failure of SAM is reasonable.

\begin{figure*}[htbp]
    \centering
    \includegraphics[width=\linewidth]{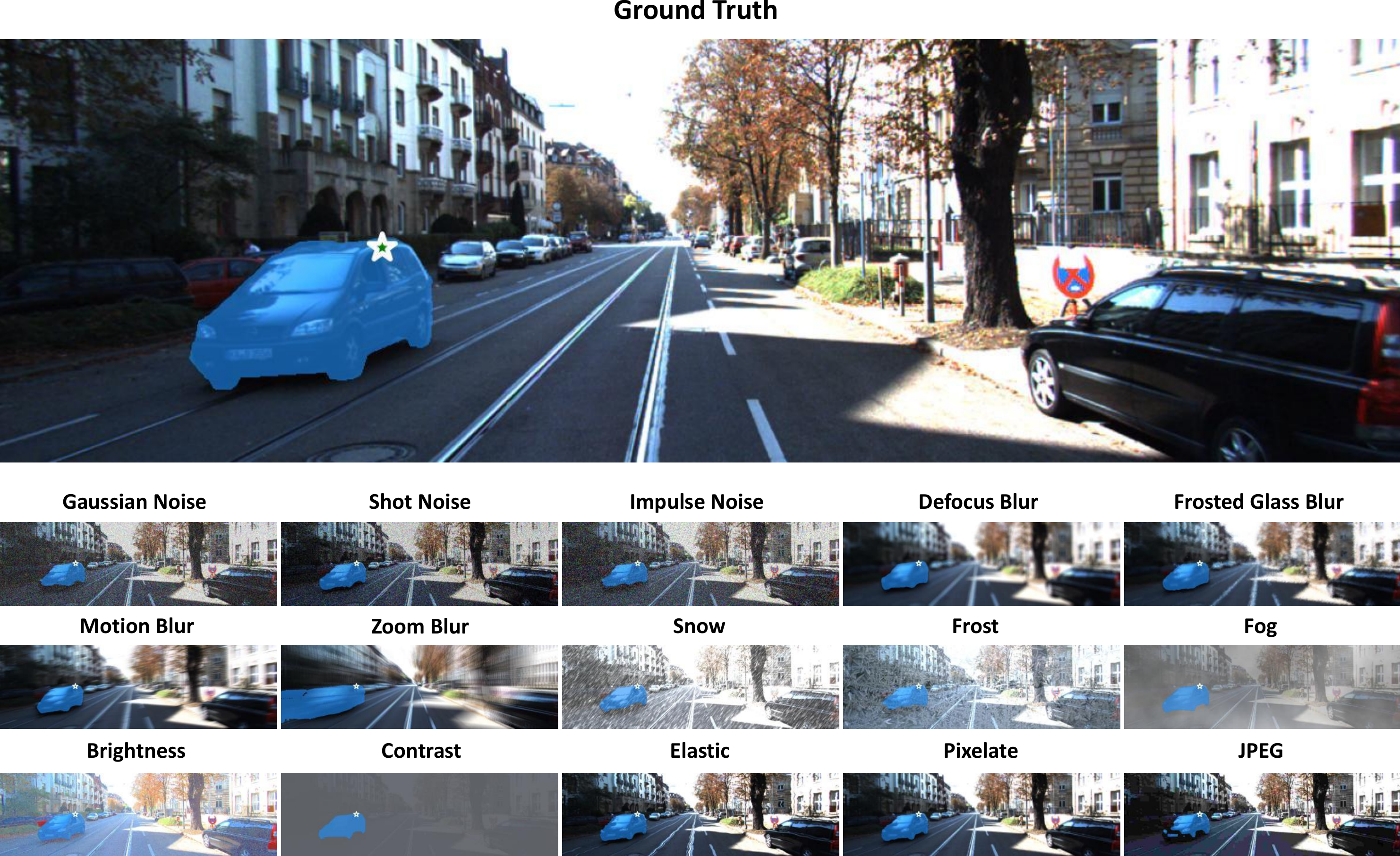}
    \caption{The visualization results on an image of KITTI. This figure shows the segmentation result of a relatively big object in the image. The segmentation results against corruption are excellent excluding zoom blur.}
    \label{fig:KITTI_result_big}
\end{figure*}

\begin{figure*}[htbp]
    \centering
    \includegraphics[width=\linewidth]{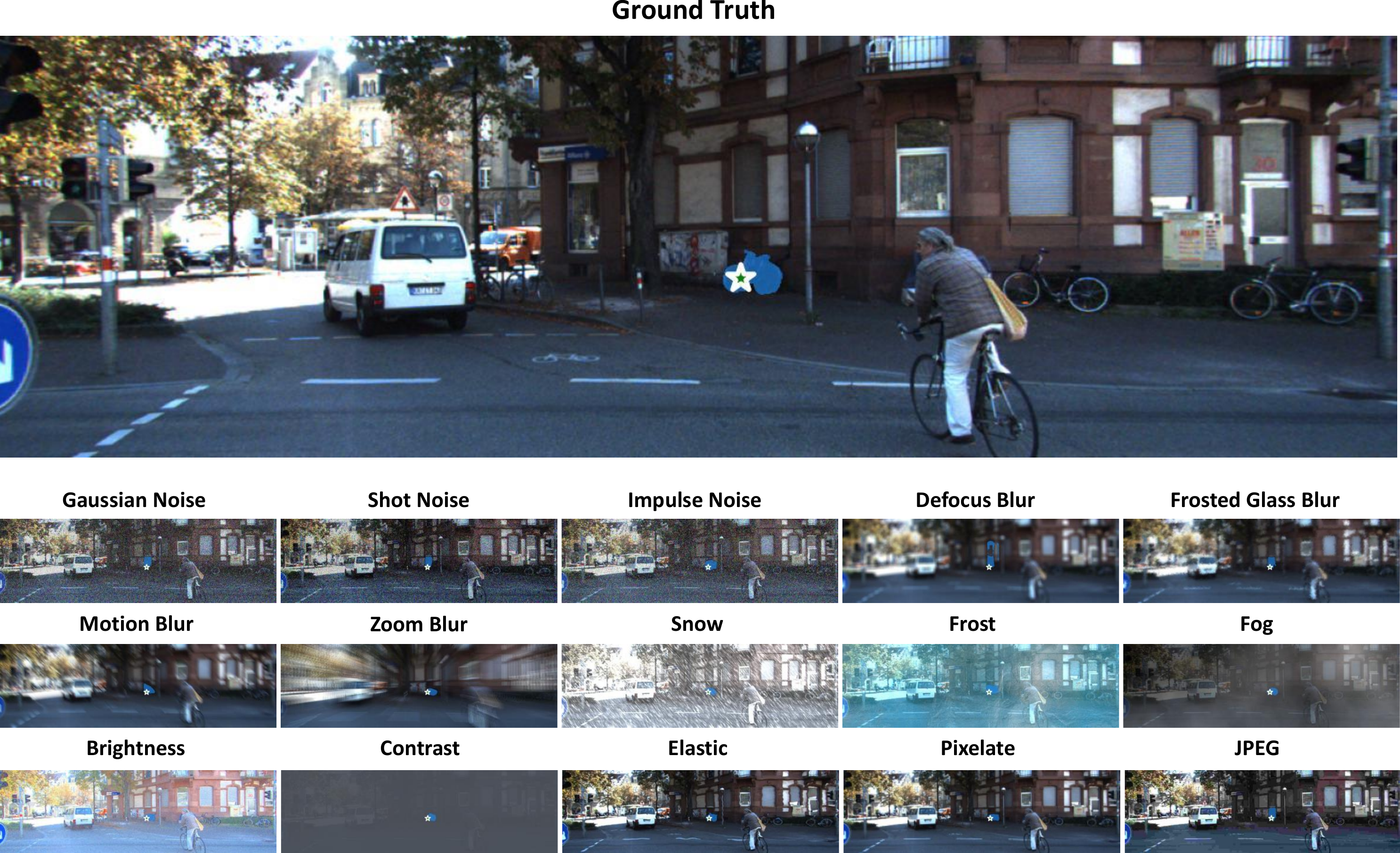}
    \caption{The visualization results on an image of KITTI. This figure shows the segmentation result of a relatively small object in the image. The segmentation results against corruption are fair.}
    \label{fig:KITTI_result_small}
\end{figure*}

\noindent\textbf{Results on KITTI against corruption.}
As shown in Fig.~\ref{fig:KITTI_result_big}, the SAM model is required to predict the segmentation result with the given point (\ie, the star in the image). We can find that SAM achieves excellent results in segmenting the car (a relatively big object on the road) under most of the corruption with the strongest severity. It only fails when facing zoom blur. As shown in Fig.~\ref{fig:KITTI_result_small}, we can find that SAM achieves fair results in segmenting the bike (a relatively small object on the road) under most of the corruption with the strongest severity. Although SAM may not accurately segment the bike, the results are close to the ground truth.

\begin{figure*}[htbp]
    \centering
    \includegraphics[width=0.8\linewidth]{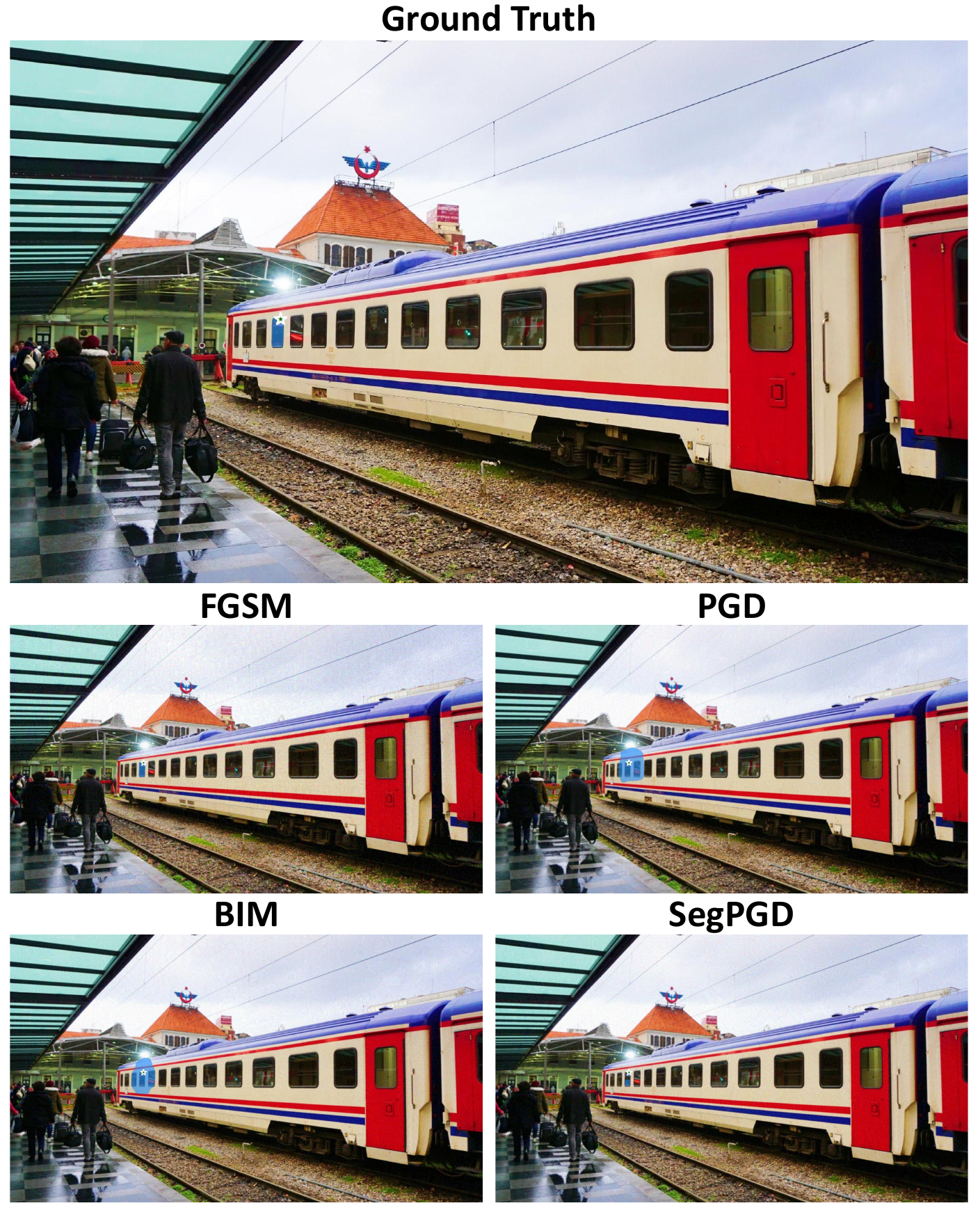}
    \caption{The visualization results of the SA-1B image under adversarial attacks. This figure displays the results on a train window. The segmentation results were not significantly affected by the FGSM attack. However, the masks under the PGD and BIM attacks exceed the original mask, while the mask under the SegPGD attack is smaller than the original mask.}
    \label{fig:adv_SAM_result}
\end{figure*}

\begin{figure*}[htbp]
    \centering
    \includegraphics[width=\linewidth]{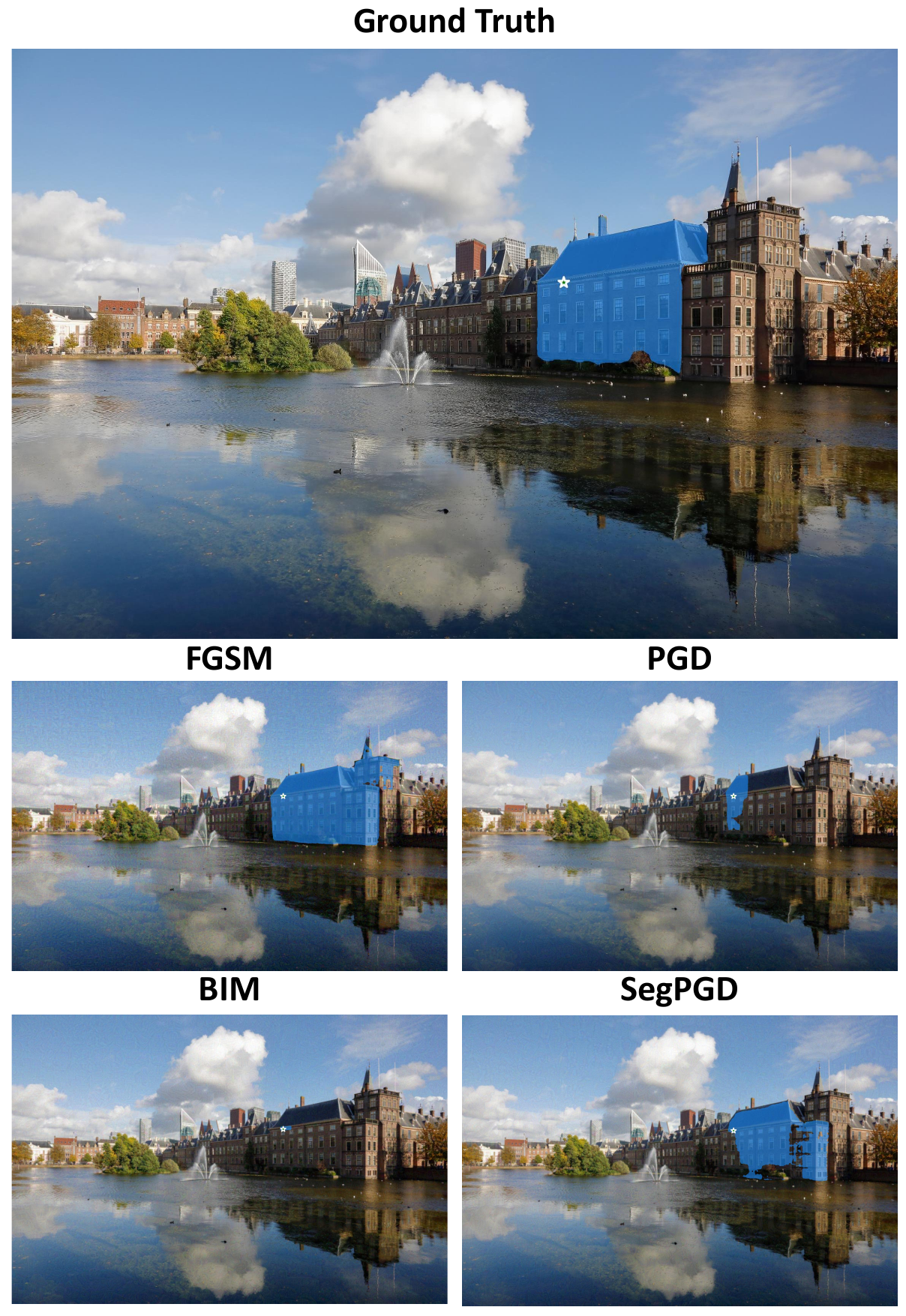}
    \caption{The visualization results of the SA-1B image under adversarial attacks. This figure displays the results on a building. The PGD and BIM attacks had a significant impact on the segmentation results.}
    \label{fig:adv_SAM2_Big_result}
\end{figure*}

\noindent\textbf{Results on SA-1B against adversarial attacks.}
In Fig.~\ref{fig:adv_SAM_result}, the SAM model is required to predict the segmentation result with the given point (\ie, the star in the image). This figure displays the results on a train window. We can observe that the segmentation results were not significantly affected by the FGSM attack. However, the masks under the PGD and BIM attacks exceed the original mask, while the mask under the SegPGD attack is smaller than the original mask. As shown in  Fig.~\ref{fig:adv_SAM2_Big_result}, we can see that SAM performs better in segmenting the building (a region with irregular edges) under FGSM and SegPGD. In contrast, the PGD and BIM attacks had a significant impact on the segmentation results, causing many parts of the building to go undetected in the mask.

\begin{figure*}[htbp]
    \centering
    \includegraphics[width=\linewidth]{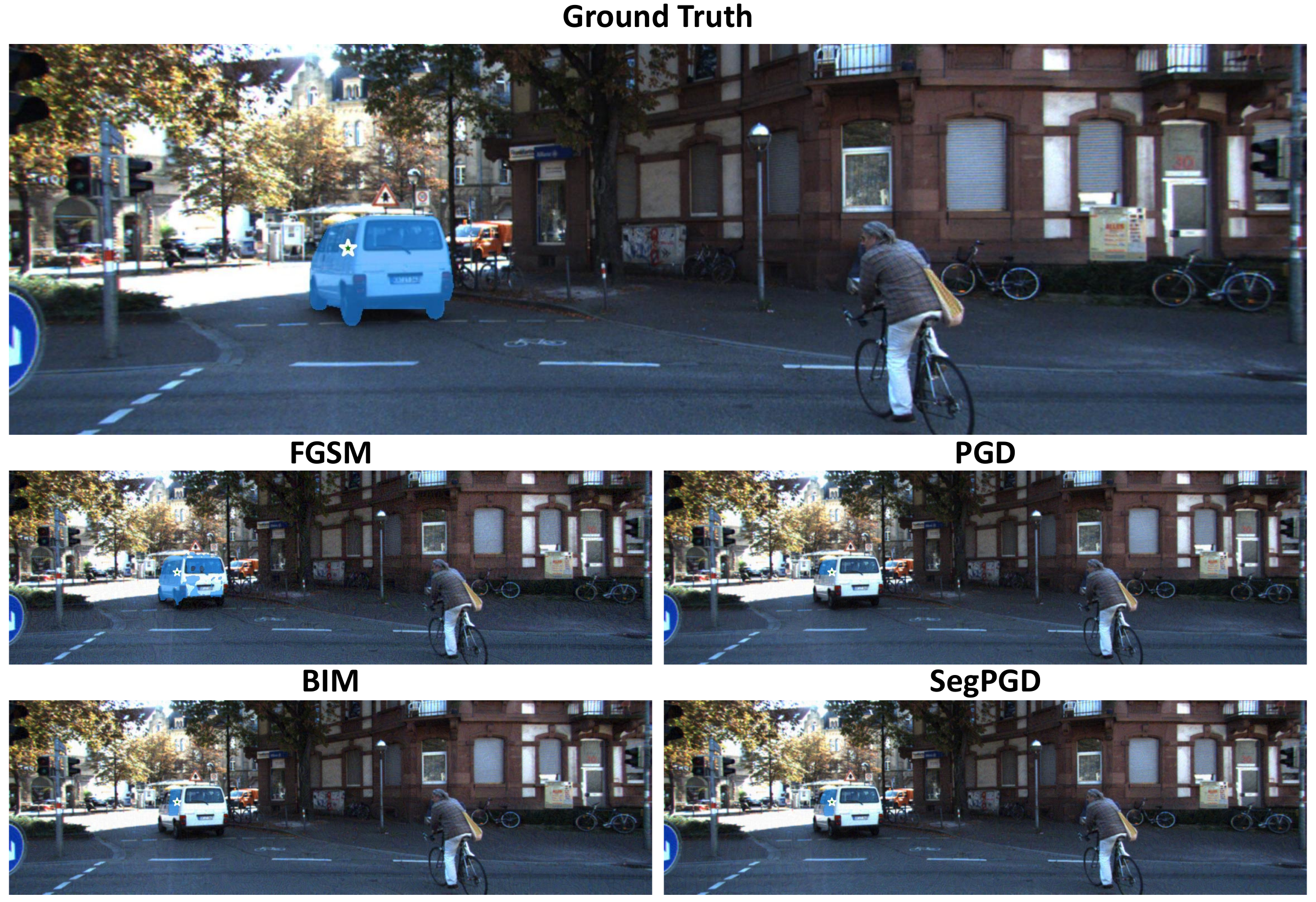}
    \caption{The visualization results on an image of KITTI. This figure shows the segmentation result of a relatively big car in the image. The segmentation results were notably influenced by the PGD and BIM attacks.}
    \label{fig:adv_KITTI_result}
\end{figure*}

\noindent\textbf{Results on KITTI against adversarial attacks.}
As shown in Fig.~\ref{fig:adv_KITTI_result}, the SAM model is required to predict the segmentation result with the given point (\ie, the star in the image). This figure shows the segmentation result of a relatively big car in the image. The segmentation results were notably influenced by the PGD, BIM, and SegPGD attacks. Conversely, the FGSM attack's influence on the results is comparatively minor. As shown in Fig.~\ref{fig:adv_KITTI_result}, this figure shows the segmentation result of a relatively small car in the image. We can observe that Sam performs well under the FGSM attack in this case, while the PGD, BIM, and SegPGD attacks all have a significant influence on this mask.

\section{Conclusion}
As the first foundation model for segmentation tasks, SAM is destined to lead the way across various tasks and in the computer vision domain. Thus the robustness evaluation on SAM is an important reference for improving the safety and reliability of foundation models. We hope the comprehensive analysis of adversarial robustness and corruption robustness on SAM can promote further study on the security of foundation models. In future work, we aim to evaluate the robustness of SAM against corruption-based adversarial attacks \cite{gao2022can,tian2021ava,guo2020watch,sun2022ala}.


\begin{table*}
\centering
\caption{Adversarial attacks results of SAM.}
\label{table:sam_adv}
\resizebox{\linewidth}{!}{
\begin{tabular}{l|llll|llll} 
\hline
                 & \multicolumn{4}{c|}{mPA Clean: 0.976232608930778}
                 & \multicolumn{4}{c}{mIoU Clean:0.9547137233317092}              \\ 
\hline
$\epsilon$ & FGSM                                    & PGD         & BIM         & SegPGD      & FGSM                                    & PGD         & BIM         & SegPGD       \\ 
\hline
0.5/255          & 0.928996319                             & 0.908836869 & 0.909020796 & 0.909854636 & 0.87347944                              & 0.850775243 & 0.849998209 & 0.879589386  \\
1/255            & 0.914257724                             & 0.890368733 & 0.888596812 & 0.890110277 & 0.849221224                             & 0.825693606 & 0.82450193  & 0.858926601  \\
2/255            & 0.903896729                             & 0.779949246 & 0.78048423  & 0.791343798 & 0.832090399                             & 0.702045795 & 0.698839971 & 0.761718543  \\
4/255            & 0.898640572                             & 0.704617135 & 0.705845363 & 0.71948928  & 0.824037058                             & 0.629457105 & 0.629236631 & 0.692308867  \\
8/255            & 0.897713578                             & 0.682245854 & 0.679297461 & 0.693165316 & 0.821877464                             & 0.607289489 & 0.602703575 & 0.665436859  \\ 
\hline
                 & \multicolumn{4}{c|}{PA background Clean: 0.999625066897009}
                 & \multicolumn{4}{c}{PA foreground Clean: 0.952840150983374}
                 \\
\hline
\end{tabular}
}
\end{table*}

\begin{table*}
\centering
\caption{Adversarial attacks results of KITTI.}
\label{table:kitti_adv}
\resizebox{\linewidth}{!}{
\begin{tabular}{l|llll|llll} 
\hline
                 & \multicolumn{4}{c|}{mPA Clean: 0.835265703465735}            
                 & \multicolumn{4}{c}{mIoU Clean:0.782404883318466}             \\ 
\hline
$\epsilon$ & FGSM                                    & PGD         & BIM         & SegPGD      & FGSM                                    & PGD         & BIM         & SegPGD       \\ 
\hline
0.5/255          & 0.796871732                             & 0.784669504 & 0.781673936 & 0.7833338   & 0.737983382                             & 0.725397844 & 0.72605909  & 0.73413779   \\
1/255            & 0.784211612                             & 0.754411781 & 0.751060767 & 0.757771992 & 0.723417764                             & 0.694846748 & 0.697139708 & 0.715124239  \\
2/255            & 0.773407942                             & 0.666288748 & 0.672777163 & 0.689335827 & 0.709875465                             & 0.621687241 & 0.620540915 & 0.650329398  \\
4/255            & 0.765117186                             & 0.621115604 & 0.622829575 & 0.633577987 & 0.700561338                             & 0.578208729 & 0.577801389 & 0.60057876   \\
8/255            & 0.767296743                             & 0.613276369 & 0.59882352  & 0.631993674 & 0.702595024                             & 0.569793901 & 0.556646505 & 0.597183662  \\ 
\hline
                 & \multicolumn{4}{c|}{PA background Clean: 0.99405102125}
                 & \multicolumn{4}{c}{PA foreground Clean: 0.676480385999999}
                 \\
\hline
\end{tabular}
}
\end{table*}

\begin{table*}
\centering
\caption{The results of adversarial attacks on big objects in KITTI.}
\label{table:kitti_big_adv}
\resizebox{\linewidth}{!}{
\begin{tabular}{l|llll|llll} 
\hline
             & \multicolumn{4}{c|}{mPA Clean: 0.812362355576453}
             & \multicolumn{4}{c}{mIoU Clean:0.776168197290638}              \\ 
\hline
$\epsilon$ & FGSM                                    & PGD         & BIM         & SegPGD      & FGSM                                    & PGD         & BIM         & SegPGD       \\ 
\hline
0.5/255      & 0.772583745                             & 0.754938745 & 0.758432951 & 0.762331041 & 0.734853013                             & 0.714879679 & 0.720446119 & 0.726477434  \\
1/255        & 0.7631925                               & 0.743660314 & 0.736570667 & 0.749761088 & 0.724534991                             & 0.70508398  & 0.699025161 & 0.713260775  \\
2/255        & 0.751274152                             & 0.674075763 & 0.671831695 & 0.685971597 & 0.710383927                             & 0.630894226 & 0.628171913 & 0.644072655  \\
4/255        & 0.741437032                             & 0.6359058   & 0.638684966 & 0.645192535 & 0.699632765                             & 0.589847213 & 0.592930934 & 0.601802078  \\
8/255        & 0.744247885                             & 0.616603806 & 0.615927569 & 0.620814379 & 0.700462402                             & 0.570293705 & 0.567166188 & 0.576312042  \\ 
\hline
             & \multicolumn{4}{c|}{PA background Clean: 0.991071857619999} 
             & \multicolumn{4}{c}{PA foreground Clean: 0.633652853644665}
             \\
\hline
\end{tabular}
}
\end{table*}

\begin{table*}
\centering
\caption{The results of adversarial attacks on small objects in KITTI.}
\label{table:kitti_small_adv}
\resizebox{\linewidth}{!}{
\begin{tabular}{l|llll|llll} 
\hline
             & \multicolumn{4}{c|}{mPA Clean: 0.83659974449138 } 
             & \multicolumn{4}{c}{mIoU Clean:0.779624332707954}             \\ 
\hline
$\epsilon$ & FGSM                                    & PGD         & BIM         & SegPGD      & FGSM                                    & PGD         & BIM         & SegPGD       \\ 
\hline
0.5/255      & 0.791476754                             & 0.77307702  & 0.773226188 & 0.777607789 & 0.729560218                             & 0.715838156 & 0.717772378 & 0.731314078  \\
1/255        & 0.779371762                             & 0.754868867 & 0.755061542 & 0.763289398 & 0.714616827                             & 0.694817631 & 0.703299456 & 0.717985119  \\
2/255        & 0.769140789                             & 0.682224103 & 0.67289554  & 0.695447042 & 0.70285925                              & 0.635710235 & 0.624906322 & 0.6721107    \\
4/255        & 0.764631703                             & 0.61141477  & 0.6085886   & 0.644434685 & 0.695676845                             & 0.577589627 & 0.579778174 & 0.625248353  \\
8/255        & 0.764046999                             & 0.589822952 & 0.58441036  & 0.622608924 & 0.692392816                             & 0.565861666 & 0.560605048 & 0.605977417  \\ 
\hline
             & \multicolumn{4}{c|}{PA background Clean: 0.999685681599999}
             & \multicolumn{4}{c}{PA foreground Clean: 0.67351380735}
             \\
\hline
\end{tabular}
}
\end{table*}

\begin{table*}
\centering
\caption{Adversarial attacks results of KITTI using MSE loss.}
\label{table:mse_kitti_adv}
\resizebox{\linewidth}{!}{
\begin{tabular}{l|llll|llll} 
\hline
             & \multicolumn{4}{c|}{mPA Clean: 0.835265703465735}
             & \multicolumn{4}{c}{mIoU Clean: 0.782404883318466}
             \\ 
\hline
$\epsilon$ & FGSM                                    & PGD         & BIM         & SegPGD      & FGSM                                    & PGD         & BIM         & SegPGD       \\ 
\hline
0.5/255      & 0.789778159                             & 0.788369437 & 0.791904967 & 0.79356285  & 0.73234129                              & 0.73081955  & 0.739074274 & 0.736241513  \\
1/255        & 0.754566262                             & 0.779166478 & 0.769769303 & 0.768693394 & 0.698888181                             & 0.724564234 & 0.716933581 & 0.708790202  \\
2/255        & 0.725861194                             & 0.682978397 & 0.694933524 & 0.683950353 & 0.670805115                             & 0.631728001 & 0.638244216 & 0.618528162  \\
4/255        & 0.710436112                             & 0.615983995 & 0.614093262 & 0.616170637 & 0.656441715                             & 0.56147345  & 0.558864013 & 0.563793216  \\
8/255        & 0.711439734                             & 0.570712045 & 0.560441902 & 0.581877081 & 0.653437313                             & 0.521950281 & 0.511362717 & 0.517012767  \\ 
\hline
             & \multicolumn{4}{c|}{PA background Clean: 0.99405102125}
             & \multicolumn{4}{c}{PA foreground Clean}
             \\
\hline
\end{tabular}
}
\end{table*}

\begin{table*}
\centering
\caption{The results of adversarial attacks on big objects in KITTI using the MSE loss.}
\label{table:mse_kitti_big_adv}
\resizebox{\linewidth}{!}{
\begin{tabular}{l|llll|llll} 
\hline
             & \multicolumn{4}{c|}{mPA Clean: 0.812362355576453}
             & \multicolumn{4}{c}{mIoU Clean: 0.776168197290638}
             \\ 
\hline
$\epsilon$ & FGSM                                    & PGD         & BIM         & SegPGD      & FGSM                                    & PGD         & BIM         & SegPGD       \\ 
\hline
0.5/255      & 0.756557804                             & 0.764936593 & 0.766952038 & 0.769215972 & 0.71600216                              & 0.723206174 & 0.726302077 & 0.728273432  \\
1/255        & 0.738249532                             & 0.747689586 & 0.746539065 & 0.735408847 & 0.697306865                             & 0.699605982 & 0.698724668 & 0.689196842  \\
2/255        & 0.719094002                             & 0.655623574 & 0.677120433 & 0.66091001  & 0.675405147                             & 0.596207224 & 0.621274131 & 0.601296882  \\
4/255        & 0.706505207                             & 0.608835782 & 0.598645095 & 0.613195288 & 0.66180238                              & 0.544081869 & 0.536971328 & 0.554474529  \\
8/255        & 0.709744996                             & 0.56097703  & 0.567986779 & 0.573959436 & 0.662147602                             & 0.498994516 & 0.504015686 & 0.507654816  \\ 
\hline
             & \multicolumn{4}{c|}{PA background Clean: 0.991071857619999}
             & \multicolumn{4}{c}{PA foreground Clean: 0.633652853644665}
             \\
\hline
\end{tabular}
}
\end{table*}

\begin{table*}
\centering
\caption{The results of adversarial attacks on small objects in KITTI using the MSE loss.}
\label{table:mse_kitti_small_adv}
\resizebox{\linewidth}{!}{
\begin{tabular}{l|llll|llll} 
\hline
             & \multicolumn{4}{c|}{mPA Clean: 0.83659974449138}
             & \multicolumn{4}{c}{mIoU Clean: 0.779624332707954}
             \\ 
\hline
$\epsilon$ & FGSM                                    & PGD         & BIM         & SegPGD      & FGSM                                    & PGD         & BIM         & SegPGD       \\ 
\hline
0.5/255      & 0.794087222                             & 0.799577576 & 0.802434374 & 0.802710534 & 0.734644878                             & 0.748415053 & 0.753653494 & 0.744838495  \\
1/255        & 0.763414763                             & 0.778465142 & 0.779729478 & 0.777914483 & 0.699760514                             & 0.725945458 & 0.723018779 & 0.717499389  \\
2/255        & 0.743997                                & 0.723579627 & 0.700984219 & 0.715249717 & 0.682373892                             & 0.675621715 & 0.653021689 & 0.654746011  \\
4/255        & 0.732621108                             & 0.627028306 & 0.620168691 & 0.638942673 & 0.669736836                             & 0.59326838  & 0.588414962 & 0.589513642  \\
8/255        & 0.7295162                               & 0.577024065 & 0.571629106 & 0.592877051 & 0.66255484                              & 0.549411719 & 0.547740013 & 0.544296306  \\ 
\hline
             & \multicolumn{4}{c|}{PA background Clean: 0.999685681599999}
             & \multicolumn{4}{c}{PA foreground Clean: 0.67351380735}
             \\
\hline
\end{tabular}
}
\end{table*}

\begin{table*}
\centering
\caption{The detailed mPA \& mIoU value of SAM on SA-1B under 15 corruptions and 5 severities.}
\resizebox{\linewidth}{!}{
\begin{tabular}{c|ccccc|ccccc}
\hline 
 & \multicolumn{5}{c|}{mPA Clean: 0.97683043 } & \multicolumn{5}{c}{mIoU Clean: 0.954812596 }\tabularnewline
\hline 
 Severity & Gaussian Noise & Shot Noise & Impulse Noise & Defocus Blur & Frosted Glass Blur & Gaussian Noise & Shot Noise & Impulse Noise & Defocus Blur & Frosted Glass Blur\tabularnewline
\hline 
1 & 0.973294958 & 0.974051937 & 0.971924972 & 0.976965344 & 0.970028863 & 0.945087461 & 0.946532199 & 0.941345115 & 0.944422714 & 0.936723961\tabularnewline
2 & 0.970645536 & 0.970640765 & 0.968576953 & 0.974763644 & 0.971788528 & 0.937493346 & 0.938221677 & 0.931315336 & 0.933845751 & 0.934074242\tabularnewline
3 & 0.965710356 & 0.967134752 & 0.965327421 & 0.968139340 & 0.951648159 & 0.922872548 & 0.927756228 & 0.922428106 & 0.905451260 & 0.885455361\tabularnewline
4 & 0.959717386 & 0.960583470 & 0.958798934 & 0.960552874 & 0.955875252 & 0.903666872 & 0.905438165 & 0.901196470 & 0.874832808 & 0.879743965\tabularnewline
5 & 0.951025024 & 0.954330475 & 0.951194722 & 0.952383592 & 0.950292203 & 0.872641085 & 0.884715711 & 0.873893057 & 0.842220740 & 0.845974755\tabularnewline
\hline 
 & Motion Blur & Zoom Blur & Snow & Frost & Fog & Motion Blur & Zoom Blur & Snow & Frost & Fog\tabularnewline
\hline 
\end{tabular}
}
\label{tab:sam_corruption_mean}
\end{table*}

\begin{table*}
\centering
\caption{The detailed foreground and background PA value of SAM on SA-1B under 15 corruptions and 5 severities.}
\resizebox{\linewidth}{!}{
\begin{tabular}{c|ccccc|ccccc}
\hline 
 & \multicolumn{5}{c|}{PA background Clean: 0.999492385 } & \multicolumn{5}{c}{PA foreground Clean: 0.954168475 }\tabularnewline
\hline 
 Severity & Gaussian Noise & Shot Noise & Impulse Noise & Defocus Blur & Frosted Glass Blur & Gaussian Noise & Shot Noise & Impulse Noise & Defocus Blur & Frosted Glass Blur\tabularnewline
\hline 
1 & 0.999263873 & 0.999079516 & 0.999115281 & 0.999348211 & 0.999344383 & 0.947326043 & 0.949024358 & 0.944734663 & 0.954582478 & 0.940713344\tabularnewline
2 & 0.998976223 & 0.998671075 & 0.998700587 & 0.999223404 & 0.999154338 & 0.942314848 & 0.942610455 & 0.938453319 & 0.950303885 & 0.944422719\tabularnewline
3 & 0.998132087 & 0.998152806 & 0.998371408 & 0.998483692 & 0.998323285 & 0.933288625 & 0.936116699 & 0.932283435 & 0.937794989 & 0.904973032\tabularnewline
4 & 0.997243656 & 0.997086999 & 0.997109717 & 0.997425641 & 0.997837220 & 0.922191117 & 0.924079942 & 0.920488150 & 0.923680107 & 0.913913284\tabularnewline
5 & 0.995332153 & 0.995623941 & 0.995456991 & 0.995966095 & 0.996056517 & 0.906717894 & 0.913037009 & 0.906932454 & 0.908801090 & 0.904527889\tabularnewline
\hline 
 & Motion Blur & Zoom Blur & Snow & Frost & Fog & Motion Blur & Zoom Blur & Snow & Frost & Fog\tabularnewline
\hline 
\end{tabular}
}
\label{tab:sam_corruption_pa_detail}
\end{table*}
\begin{table*}
\centering
\caption{The detailed foreground and background IoU value of SAM on SA-1B under 15 corruptions and 5 severities.}
\resizebox{\linewidth}{!}{
\begin{tabular}{c|ccccc|ccccc}
\hline 
 & \multicolumn{5}{c|}{IoU background Clean: 0.998546789 } & \multicolumn{5}{c}{IoU foreground Clean: 0.911078402 }\tabularnewline
\hline 
 Severity & Gaussian Noise & Shot Noise & Impulse Noise & Defocus Blur & Frosted Glass Blur & Gaussian Noise & Shot Noise & Impulse Noise & Defocus Blur & Frosted Glass Blur\tabularnewline
\hline 
1 & 0.998242620 & 0.998170352 & 0.998083348 & 0.998456196 & 0.998440071 & 0.891932302 & 0.894894047 & 0.884606882 & 0.890389232 & 0.875007850\tabularnewline
2 & 0.997974458 & 0.997662629 & 0.997602673 & 0.998343203 & 0.998316403 & 0.877012234 & 0.878780725 & 0.865027998 & 0.869348300 & 0.869832080\tabularnewline
3 & 0.996992530 & 0.997089753 & 0.997225656 & 0.997599993 & 0.997382879 & 0.848752565 & 0.858422703 & 0.847630555 & 0.813302527 & 0.773527842\tabularnewline
4 & 0.996030086 & 0.995860380 & 0.995940750 & 0.996439172 & 0.996888559 & 0.811303658 & 0.815015950 & 0.806452190 & 0.753226444 & 0.762599370\tabularnewline
5 & 0.993970403 & 0.994292427 & 0.994066089 & 0.994911929 & 0.994984654 & 0.751311766 & 0.775138995 & 0.753720026 & 0.689529551 & 0.696964857\tabularnewline
\hline 
 & Motion Blur & Zoom Blur & Snow & Frost & Fog & Motion Blur & Zoom Blur & Snow & Frost & Fog\tabularnewline
\hline 
\end{tabular}
}
\label{tab:sam_corruption_iou_detail}
\end{table*}

\begin{table*}
\centering
\caption{The detailed mPA \& mIoU value of SAM on KITTI under 15 corruptions and 5 severities.}
\resizebox{\linewidth}{!}{
\begin{tabular}{c|ccccc|ccccc}
\hline 
 & \multicolumn{5}{c|}{mPA Clean: 0.821998731 } & \multicolumn{5}{c}{mIoU Clean: 0.773630113 }\tabularnewline
\hline 
 Severity & Gaussian Noise & Shot Noise & Impulse Noise & Defocus Blur & Frosted Glass Blur & Gaussian Noise & Shot Noise & Impulse Noise & Defocus Blur & Frosted Glass Blur\tabularnewline
\hline 
1 & 0.827509692  & 0.826566758 & 0.834727599  & 0.834975945  & 0.837122162  & 0.767954580  & 0.768185019 & 0.770431247  & 0.764663405  & 0.769081643 \tabularnewline
2 & 0.826017267  & 0.826535321 & 0.822316130  & 0.832771794  & 0.843622906  & 0.760825995  & 0.763999900 & 0.757401185  & 0.755053715  & 0.763684937 \tabularnewline
3 & 0.827090490  & 0.826992034 & 0.825577821  & 0.826367274  & 0.822930303  & 0.752743924  & 0.763680784 & 0.753456891  & 0.736631417  & 0.721030296 \tabularnewline
4 & 0.829997510  & 0.829871028 & 0.829574901  & 0.825402446  & 0.828452159  & 0.744140436  & 0.753289627 & 0.744088001  & 0.715587985  & 0.713900905 \tabularnewline
5 & 0.825246450  & 0.828853238 & 0.830706907  & 0.820415930  & 0.824765924  & 0.725069595  & 0.739610783 & 0.727930224  & 0.691989424  & 0.688195978 \tabularnewline
\hline 
 & Motion Blur & Zoom Blur & Snow & Frost & Fog & Motion Blur & Zoom Blur & Snow & Frost & Fog\tabularnewline
\hline 
\end{tabular}
}
\label{tab:kitti_corruption_mean}
\end{table*}
\begin{table*}
\centering
\caption{The detailed foreground and background PA value of SAM on KITTI under 15 corruptions and 5 severities.}
\resizebox{\linewidth}{!}{
\begin{tabular}{c|ccccc|ccccc}
\hline 
 & \multicolumn{5}{c|}{PA background Clean: 0.998031105 } & \multicolumn{5}{c}{PA foreground Clean: 0.645966358 }\tabularnewline
\hline 
 Severity & Gaussian Noise & Shot Noise & Impulse Noise & Defocus Blur & Frosted Glass Blur & Gaussian Noise & Shot Noise & Impulse Noise & Defocus Blur & Frosted Glass Blur\tabularnewline
\hline 
1 & 0.996753439 & 0.996981631 & 0.996297141 & 0.996662272 & 0.997359415 & 0.658265944 & 0.656151884 & 0.673158056 & 0.673289619 & 0.676884909\tabularnewline
2 & 0.996201991 & 0.995351684 & 0.996110646 & 0.996485383 & 0.996356449 & 0.655832543 & 0.657718959 & 0.648521616 & 0.669058205 & 0.690889363\tabularnewline
3 & 0.994656792 & 0.996076379 & 0.995445121 & 0.996402052 & 0.994740247 & 0.659524188 & 0.657907690 & 0.655710521 & 0.656332497 & 0.651120359\tabularnewline
4 & 0.991481828 & 0.993285678 & 0.992413799 & 0.990837446 & 0.989766433 & 0.668513193 & 0.666456378 & 0.666736002 & 0.659967445 & 0.667137885\tabularnewline
5 & 0.987493336 & 0.989886908 & 0.986873163 & 0.985586127 & 0.987335106 & 0.662999565 & 0.667819569 & 0.674540651 & 0.655245734 & 0.662196742\tabularnewline
\hline 
 & Motion Blur & Zoom Blur & Snow & Frost & Fog & Motion Blur & Zoom Blur & Snow & Frost & Fog\tabularnewline
\hline 
\end{tabular}
}
\label{tab:kitti_corruption_pa_detail}
\end{table*}
\begin{table*}
\centering
\caption{The detailed foreground and background IoU value of SAM on KITTI under 15 corruptions and 5 severities.}
\resizebox{\linewidth}{!}{
\begin{tabular}{c|ccccc|ccccc}
\hline 
 & \multicolumn{5}{c|}{IoU background Clean: 0.978571902 } & \multicolumn{5}{c}{IoU foreground Clean: 0.568688324 }\tabularnewline
\hline 
 Severity & Gaussian Noise & Shot Noise & Impulse Noise & Defocus Blur & Frosted Glass Blur & Gaussian Noise & Shot Noise & Impulse Noise & Defocus Blur & Frosted Glass Blur\tabularnewline
\hline 
1 & 0.977692697 & 0.978238207 & 0.977379872 & 0.978038252 & 0.978903832 & 0.558216464 & 0.558131831 & 0.563482622 & 0.551288559 & 0.559259455\tabularnewline
2 & 0.977130259 & 0.976290998 & 0.977174565 & 0.977841447 & 0.978041911 & 0.544521730 & 0.551708803 & 0.537627804 & 0.532265981 & 0.549327963\tabularnewline
3 & 0.976706324 & 0.978208188 & 0.976868855 & 0.978672733 & 0.976151859 & 0.528781524 & 0.549153380 & 0.530044926 & 0.494590101 & 0.465908733\tabularnewline
4 & 0.974076829 & 0.975139507 & 0.974122214 & 0.973197392 & 0.971786010 & 0.514204042 & 0.531439748 & 0.514053788 & 0.457978578 & 0.456015799\tabularnewline
5 & 0.971069568 & 0.972343777 & 0.970251885 & 0.967901867 & 0.969592797 & 0.479069623 & 0.506877789 & 0.485608562 & 0.416076981 & 0.406799158\tabularnewline
\hline 
 & Motion Blur & Zoom Blur & Snow & Frost & Fog & Motion Blur & Zoom Blur & Snow & Frost & Fog\tabularnewline
\hline 
\end{tabular}
}
\label{tab:kitti_corruption_iou_detail}
\end{table*}

\begin{table*}
\centering
\caption{The detailed mPA \& mIoU value of SAM on big objects of KITTI under 15 corruptions and 5 severities.}
\resizebox{\linewidth}{!}{
\begin{tabular}{c|ccccc|ccccc}
\hline 
 & \multicolumn{5}{c|}{mPA Clean: 0.817688259 } & \multicolumn{5}{c}{mIoU Clean: 0.776732011 }\tabularnewline
\hline 
 Severity & Gaussian Noise & Shot Noise & Impulse Noise & Defocus Blur & Frosted Glass Blur & Gaussian Noise & Shot Noise & Impulse Noise & Defocus Blur & Frosted Glass Blur\tabularnewline
\hline 
1 & 0.824544634 & 0.823148437 & 0.822386910 & 0.827915474 & 0.823614760 & 0.777264717 & 0.779445930 & 0.773900195 & 0.778729022 & 0.780460438\tabularnewline
2 & 0.819575671 & 0.823762124 & 0.822097876 & 0.828088453 & 0.823572151 & 0.771742408 & 0.775026699 & 0.770215948 & 0.777153217 & 0.776091159\tabularnewline
3 & 0.818345612 & 0.821489833 & 0.816559489 & 0.826272027 & 0.819788413 & 0.762366223 & 0.769154593 & 0.761533562 & 0.765849686 & 0.759751372\tabularnewline
4 & 0.815737855 & 0.822831050 & 0.816747346 & 0.819886683 & 0.824792242 & 0.752831653 & 0.762779204 & 0.751869062 & 0.753275214 & 0.758675952\tabularnewline
5 & 0.814497068 & 0.820328600 & 0.816744708 & 0.813016249 & 0.821662629 & 0.736033262 & 0.749445040 & 0.736320497 & 0.740147459 & 0.744921491\tabularnewline
\hline 
 & Motion Blur & Zoom Blur & Snow & Frost & Fog & Motion Blur & Zoom Blur & Snow & Frost & Fog\tabularnewline
\hline 
\end{tabular}
}
\label{tab:kitti_corruption_big_mask_mean}
\end{table*}
\begin{table*}
\centering
\caption{The detailed foreground and background PA value of SAM on big objects of KITTI under 15 corruptions and 5 severities.}
\resizebox{\linewidth}{!}{
\begin{tabular}{c|ccccc|ccccc}
\hline 
 Severity & \multicolumn{5}{c|}{PA background Clean: 0.991817059 } & \multicolumn{5}{c}{PA foreground Clean: 0.643559458 }\tabularnewline
\hline 
 & Gaussian Noise & Shot Noise & Impulse Noise & Defocus Blur & Frosted Glass Blur & Gaussian Noise & Shot Noise & Impulse Noise & Defocus Blur & Frosted Glass Blur\tabularnewline
\hline 
1 & 0.990301260 & 0.990281745 & 0.989882451 & 0.991167381 & 0.992253791 & 0.658788009 & 0.656015128 & 0.654891369 & 0.664663566 & 0.654975729\tabularnewline
2 & 0.989247489 & 0.988486916 & 0.984404303 & 0.990608807 & 0.990738046 & 0.649903853 & 0.659037332 & 0.659791449 & 0.665568099 & 0.656406256\tabularnewline
3 & 0.981644931 & 0.989466848 & 0.982586795 & 0.985630523 & 0.987692371 & 0.655046292 & 0.653512817 & 0.650532182 & 0.666913532 & 0.651884455\tabularnewline
4 & 0.975256524 & 0.982440456 & 0.977639478 & 0.983497569 & 0.985965236 & 0.656219187 & 0.663221644 & 0.655855213 & 0.656275797 & 0.663619249\tabularnewline
5 & 0.966347780 & 0.978000803 & 0.967255013 & 0.982396333 & 0.979526999 & 0.662646357 & 0.662656396 & 0.666234403 & 0.643636165 & 0.663798258\tabularnewline
\hline 
 & Motion Blur & Zoom Blur & Snow & Frost & Fog & Motion Blur & Zoom Blur & Snow & Frost & Fog\tabularnewline
\hline 
\end{tabular}
}
\label{tab:kitti_corruption_big_mask_pa_detail}
\end{table*}
\begin{table*}
\centering
\caption{The detailed foreground and background IoU value of SAM on big objects of KITTI under 15 corruptions and 5 severities.}
\resizebox{\linewidth}{!}{
\begin{tabular}{c|ccccc|ccccc}
\hline 
 & \multicolumn{5}{c|}{IoU background Clean: 0.950995222 } & \multicolumn{5}{c}{IoU foreground Clean: 0.602468801 }\tabularnewline
\hline 
 Severity & Gaussian Noise & Shot Noise & Impulse Noise & Defocus Blur & Frosted Glass Blur & Gaussian Noise & Shot Noise & Impulse Noise & Defocus Blur & Frosted Glass Blur\tabularnewline
\hline 
1 & 0.951123078 & 0.951172114 & 0.950142585 & 0.950374245 & 0.953057698 & 0.603406357 & 0.607719747 & 0.597657805 & 0.607083799 & 0.607863178\tabularnewline
2 & 0.949593559 & 0.949717430 & 0.945390965 & 0.949993699 & 0.951336971 & 0.593891257 & 0.600335968 & 0.595040932 & 0.604312735 & 0.600845347\tabularnewline
3 & 0.940864374 & 0.947968338 & 0.941614536 & 0.947216319 & 0.949522285 & 0.583868072 & 0.590340848 & 0.581452588 & 0.584483054 & 0.569980458\tabularnewline
4 & 0.937129139 & 0.944515930 & 0.938975861 & 0.945785643 & 0.947024836 & 0.568534166 & 0.581042478 & 0.564762262 & 0.560764786 & 0.570327069\tabularnewline
5 & 0.930268006 & 0.941484607 & 0.930506677 & 0.943389483 & 0.942394835 & 0.541798518 & 0.557405475 & 0.542134318 & 0.536905434 & 0.547448146\tabularnewline
\hline 
 & Motion Blur & Zoom Blur & Snow & Frost & Fog & Motion Blur & Zoom Blur & Snow & Frost & Fog\tabularnewline
\hline 
\end{tabular}
}
\label{tab:kitti_corruption_big_mask_iou_detail}
\end{table*}

\begin{table*}
\centering
\caption{The detailed mPA \& mIoU value of SAM on small objects of KITTI under 15 corruptions and 5 severities.}
\resizebox{\linewidth}{!}{
\begin{tabular}{c|ccccc|ccccc}
\hline 
 & \multicolumn{5}{c|}{mPA Clean: 0.845691467 } & \multicolumn{5}{c}{mIoU Clean: 0.782262591 }\tabularnewline
\hline 
 Severity & Gaussian Noise & Shot Noise & Impulse Noise & Defocus Blur & Frosted Glass Blur & Gaussian Noise & Shot Noise & Impulse Noise & Defocus Blur & Frosted Glass Blur\tabularnewline
\hline 
1 & 0.850790394 & 0.858883705 & 0.854653978 & 0.860443176 & 0.857061718 & 0.774985797 & 0.780847891 & 0.777112329 & 0.757984746 & 0.767346301\tabularnewline
2 & 0.850245101 & 0.852492774 & 0.853715358 & 0.852120218 & 0.861124392 & 0.771048960 & 0.773252549 & 0.767994066 & 0.731055425 & 0.746718308\tabularnewline
3 & 0.848452539 & 0.847916763 & 0.847278064 & 0.852352957 & 0.847477582 & 0.760131763 & 0.763885823 & 0.752534684 & 0.688034408 & 0.681690461\tabularnewline
4 & 0.842638482 & 0.843449371 & 0.846432726 & 0.843618019 & 0.850695786 & 0.737793441 & 0.743000344 & 0.739779126 & 0.651111302 & 0.664060894\tabularnewline
5 & 0.837899002 & 0.843256355 & 0.832528260 & 0.836772886 & 0.845213988 & 0.714934054 & 0.726434794 & 0.708943708 & 0.626129961 & 0.627538398\tabularnewline
\hline 
 & Motion Blur & Zoom Blur & Snow & Frost & Fog & Motion Blur & Zoom Blur & Snow & Frost & Fog\tabularnewline
\hline 
\end{tabular}
}
\label{tab:kitti_corruption_small_mask_mean}
\end{table*}
\begin{table*}
\centering
\caption{The detailed foreground and background PA value of SAM on small objects of KITTI under 15 corruptions and 5 severities.}
\resizebox{\linewidth}{!}{
\begin{tabular}{c|ccccc|ccccc}
\hline 
 & \multicolumn{5}{c|}{PA background Clean: 0.999526278 } & \multicolumn{5}{c}{PA foreground Clean: 0.691856657 }\tabularnewline
\hline 
 Severity & Gaussian Noise & Shot Noise & Impulse Noise & Defocus Blur & Frosted Glass Blur & Gaussian Noise & Shot Noise & Impulse Noise & Defocus Blur & Frosted Glass Blur\tabularnewline
\hline 
1 & 0.999152714 & 0.998567927 & 0.998902985 & 0.999222958 & 0.999332330 & 0.702428074 & 0.719199482 & 0.710404970 & 0.721663395 & 0.714791105\tabularnewline
2 & 0.997865891 & 0.998585835 & 0.998462137 & 0.998457265 & 0.998692020 & 0.702624310 & 0.706399713 & 0.708968580 & 0.705783172 & 0.723556765\tabularnewline
3 & 0.998217274 & 0.998059935 & 0.996252307 & 0.996920662 & 0.996806880 & 0.698687805 & 0.697773590 & 0.698303820 & 0.707785251 & 0.698148283\tabularnewline
4 & 0.996876972 & 0.997324193 & 0.995665779 & 0.995076974 & 0.994793233 & 0.688399992 & 0.689574548 & 0.697199672 & 0.692159064 & 0.706598339\tabularnewline
5 & 0.995394015 & 0.996372267 & 0.994189389 & 0.990873175 & 0.992019941 & 0.680403990 & 0.690140443 & 0.670867132 & 0.682672597 & 0.698408034\tabularnewline
\hline 
 & Motion Blur & Zoom Blur & Snow & Frost & Fog & Motion Blur & Zoom Blur & Snow & Frost & Fog\tabularnewline
\hline 
\end{tabular}
}
\label{tab:kitti_corruption_small_mask_pa_detail}
\end{table*}
\begin{table*}
\centering
\caption{The detailed foreground and background IoU value of SAM on small objects of KITTI under 15 corruptions and 5 severities.}
\resizebox{\linewidth}{!}{
\begin{tabular}{c|ccccc|ccccc}
\hline 
 & \multicolumn{5}{c|}{IoU background Clean: 0.99832729 } & \multicolumn{5}{c}{IoU foreground Clean: 0.566197893 }\tabularnewline
\hline 
 Severity & Gaussian Noise & Shot Noise & Impulse Noise & Defocus Blur & Frosted Glass Blur & Gaussian Noise & Shot Noise & Impulse Noise & Defocus Blur & Frosted Glass Blur\tabularnewline
\hline 
1 & 0.997961808 & 0.997381065 & 0.997711180 & 0.998053463 & 0.998146426 & 0.552009786 & 0.564314717 & 0.556513477 & 0.517916030 & 0.536546176\tabularnewline
2 & 0.996712313 & 0.997434754 & 0.997283124 & 0.997280293 & 0.997565149 & 0.545385607 & 0.549070343 & 0.538705009 & 0.464830556 & 0.495871469\tabularnewline
3 & 0.997040189 & 0.996855867 & 0.995090242 & 0.995725470 & 0.995683530 & 0.523223337 & 0.530915778 & 0.509979126 & 0.380343345 & 0.367697392\tabularnewline
4 & 0.995688711 & 0.996163298 & 0.994525220 & 0.993899876 & 0.993672554 & 0.479898172 & 0.489837389 & 0.485033031 & 0.308322728 & 0.334449234\tabularnewline
5 & 0.994249437 & 0.995170064 & 0.993049768 & 0.989676587 & 0.990887362 & 0.435618672 & 0.457699524 & 0.424837648 & 0.262583334 & 0.264189434\tabularnewline
\hline 
 & Motion Blur & Zoom Blur & Snow & Frost & Fog & Motion Blur & Zoom Blur & Snow & Frost & Fog\tabularnewline
\hline 
\end{tabular}
}
\label{tab:kitti_corruption_small_mask_iou_detail}
\end{table*}

\begin{figure*}[htbp]
    \centering
    \includegraphics[width=\linewidth]{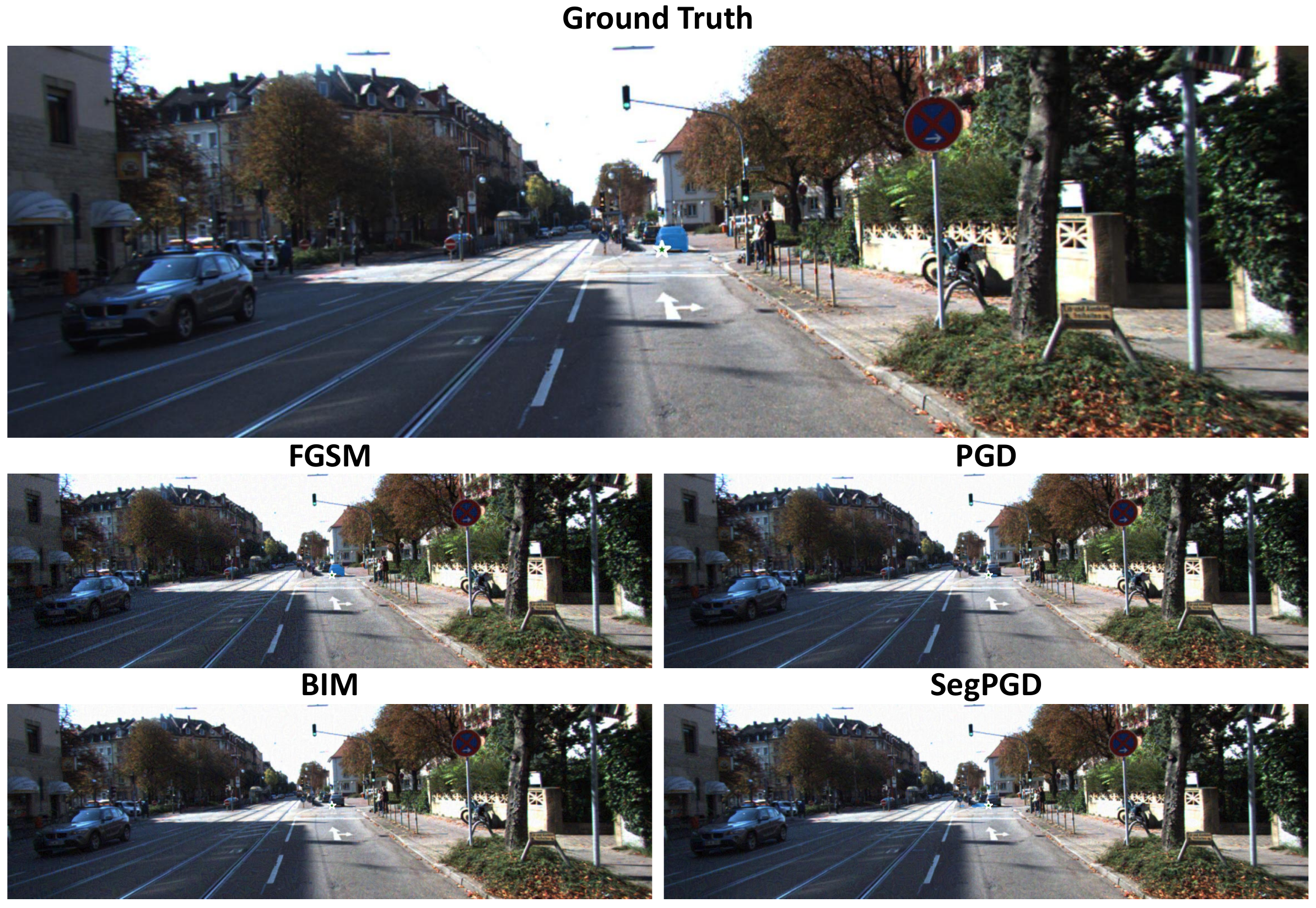}
    \caption{The visualization results on an image of KITTI. This figure shows the segmentation result of a relatively small car in the image. PGD, BIM, and SegPGD attacks all have a considerable influence on this mask.}
    \label{fig:adv_KITTI_Small_result}
\end{figure*}


\clearpage
\newpage
\balance
\bibliographystyle{IEEEtran}
\bibliography{Citation}

\end{document}